\documentclass{article} 
\usepackage{iclr2026_paper,times}

\usepackage{amsmath,amsfonts,bm}

\def\eqref#1{equation~\ref{#1}}

\def\1{\bm{1}}

\DeclareMathAlphabet{\mathsfit}{\encodingdefault}{\sfdefault}{m}{sl}
\SetMathAlphabet{\mathsfit}{bold}{\encodingdefault}{\sfdefault}{bx}{n}

\usepackage{url}

\usepackage{xcolor}         
\usepackage{tabularx, multirow, multicol, graphicx, makecell, enumitem, color, soul, colortbl, booktabs, amssymb}
\usepackage{floatrow}
\newfloatcommand{capbtabbox}{table}[][\FBwidth]

\usepackage{algorithm}
\usepackage{algpseudocode} 

\usepackage{siunitx}
\usepackage{sourcecodepro}
\usepackage[most]{tcolorbox}
\usepackage{newfloat}
\usepackage{listings}
\DeclareCaptionStyle{ruled}{labelfont=normalfont,labelsep=colon,strut=off}
\definecolor{linenum}{RGB}{140,140,140}
\floatstyle{ruled}
\newfloat{listing}{tb}{lst}{}
\floatname{listing}{Listing}

\usepackage{xspace}

\usepackage{wrapfig}  
\usepackage{caption}

\definecolor{greycolor}{RGB}{105,105,105}

\newcommand{\red}[1]{\textcolor{googlered}{#1}}

\definecolor{vscodegreen}{RGB}{16,185,129} 
\definecolor{lightgreen}{RGB}{240, 251, 237} 
\definecolor{lightred}{RGB}{251, 240, 237} 

\definecolor{googleblue}{HTML}{4285F4}
\definecolor{googlered}{HTML}{EA4335}
\definecolor{googleyellow}{HTML}{FBBC05}
\definecolor{googlegreen}{HTML}{34A853}
\definecolor{googlepurple}{HTML}{A142F4}
\definecolor{annocolor}{RGB}{133,112,255}

\newcommand{\MODEL}{Code2Video\xspace}
\newcommand{\our}{Code2Video\xspace}

\newcommand{\eg}[1]{\textit{e.g.,}}
\newcommand{\ie}[1]{\textit{i.e.,}}
\newcommand{\quiz}{TeachQuiz\xspace}

\newcommand{\DATASET}{MMMC\xspace}

\definecolor{citecolor}{HTML}{0071bc}
\usepackage{hyperref}
\hypersetup{breaklinks=true,colorlinks,citecolor=citecolor,urlcolor=googleblue}

\usepackage{cleveref, afterpage}

\title{%
  \includegraphics[width=0.55cm, height=0.55cm]{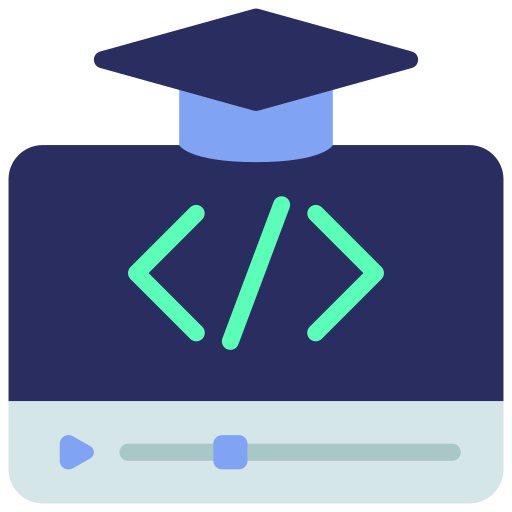}%
  \hspace{0.3em}
  \MODEL: A Code-centric Paradigm for \\ Educational Video Generation
}

\usepackage{marvosym}

\author{Yanzhe Chen\textsuperscript{*}
    \quad
    Kevin Qinghong Lin\textsuperscript{*}
    \quad
    Mike Zheng Shou\textsuperscript{\Letter}
    \\[1.2ex]
    Show Lab, National University of Singapore
    \\[1.2ex]
    \url{https://showlab.github.io/Code2Video/}
}

\iclrfinalcopy

\begin{document}

\maketitle

\blfootnote{\textsuperscript{*}Equal contribution \quad \textsuperscript{\Letter}Corresponding author: mike.zheng.shou@gmail.com}

\vspace{-1.2cm}
\begin{abstract}
\vspace{-0.1cm}
While recent generative models advance pixel-space video synthesis, they remain limited in producing professional educational videos, which demand disciplinary knowledge, precise visual structures, and coherent transitions, limiting their applicability in educational scenarios.
Intuitively, such requirements are better addressed through the manipulation of a renderable environment, which can be explicitly controlled via logical commands (\eg, code).
In this work, we propose~\textbf{\our}, a code-centric agent framework for generating educational videos via executable Python code. 
The framework comprises three collaborative agents: \textit{{(\romannumeral1) Planner}}, which structures lecture content into temporally coherent flows and prepares corresponding visual assets; \textit{{(\romannumeral2) Coder}}, which converts structured instructions into executable Python codes while incorporating scope-guided auto-fix to enhance efficiency; 
and \textit{{(\romannumeral3) Critic}}, which leverages vision-language models (VLM) with visual anchor prompts to refine spatial layout and ensure clarity.
To support systematic evaluation, we build \textbf{\DATASET}, a benchmark of professionally produced, discipline-specific educational videos. We evaluate \DATASET across diverse dimensions, including VLM-as-a-Judge aesthetic scores, code efficiency, and particularly, \textbf{\quiz}, 
a novel end-to-end metric that quantifies how well a VLM, after unlearning, can recover knowledge by watching the generated videos.
Our results demonstrate the potential of \MODEL as a scalable, interpretable, and controllable approach, achieving 40\% improvement over direct code generation and producing videos comparable to human-crafted tutorials.
The code and datasets are available at \href{https://github.com/showlab/Code2Video}{https://github.com/showlab/Code2Video}.

\end{abstract}


\section{Introduction}

\vspace{-0.2cm}

\quad\textit{“If you want to master something, teach it.”} -- Richard Feynman

Recent advances in natural video generation have made remarkable progress in \textit{pixel} space. End-to-end solutions, including diffusion-based~\citep{ho2022imagen, weng2024genrec} and autoregressive architectures~\citep{weng2024art, yuan2025lumos}, can synthesize visually compelling videos directly from text prompts (\ie~\textbf{Text2Video}), achieving fine appearance and short-form fidelity. Yet these models struggle when the task requires long-form reasoning or multi-entity interaction~\citep{li2024survey}. To overcome these limitations, recent works have moved toward multi-agent pipelines, where complex video generation is decomposed into collaborative subtasks, allowing iterative refinement, temporal structuring~\citep{yuan2024mora, huang2024genmac, xie2024dreamfactory}.

Educational videos that aim to teach subject-specific knowledge face unique challenges in the reasoning era.
Unlike short-form entertainment, educational content must integrate deep domain expertise~\citep{clark2023learning}, carefully designed animations or transitions, and step-by-step reasoning~\citep{bao2009learning} to support actual skill acquisition. This raises two fundamental challenges: \textbf{(\romannumeral1)} How to create high-quality educational videos that maintain both temporal coherence—concepts introduced, expanded, and reinforced in logical sequence—and spatial clarity—elements arranged legibly without occlusion; and \textbf{(\romannumeral2)} How to evaluate educational videos beyond appearance, ensuring that they are educationally effective and semantically aligned with the intended learning topic. 
Existing video generation pipelines rarely satisfy these requirements, leaving a critical gap for agentic methods that unify temporal planning, spatial organization, and educational assessment.

We are motivated by the intuition that code provides a uniquely suitable substrate for educational video generation. Unlike black-box models, code-centric pipelines are \textit{scalable}, since new visualizations and external assets can be modularly integrated; \textit{interpretable}, as every sequence, layout, and rendering decision is explicitly scripted and thus auditable; and \textit{controllable}, enabling precise temporal sequencing and spatial organization through programmatic specification.

\begin{figure}[!t]
  \centering
  \includegraphics[width=1.0\linewidth]{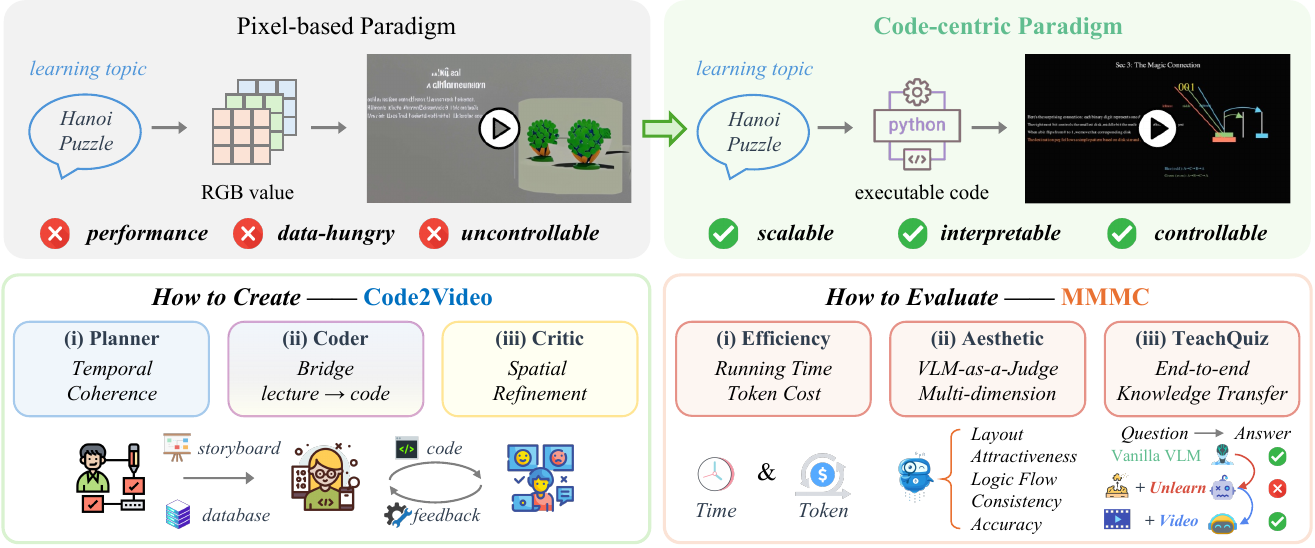}
  \vspace{-0.4cm}
  \caption{Overview of \textbf{\MODEL}. A code-centric paradigm for educational video generation, where Planner ensures temporal flow, Coder bridges instructions to executable animations, and Critic refines spatial layout. Evaluation is performed on \textbf{\DATASET} with multi-dimensional metrics.}
  \label{qiancai}
\end{figure}

Building on these insights, we propose \textbf{\MODEL}, an agentic, code-centric framework for generating high-quality educational videos. The system decomposes the task into three agents: the \textit{Planner} sequences concepts, examples, and recaps into a coherent lecture flow; the \textit{Coder} translates structured instructions into executable Manim code, yielding precise, editable visualizations with consistent layout and timing; and the \textit{Critic} leverages multimodal feedback and visual anchor prompts to refine spatial organization and ensure alignment with learning objectives.
This tri-agent design explicitly models the temporal and spatial structure of instruction, while grounding the entire pipeline in transparent, reproducible, and extensible code.

To evaluate this paradigm, we propose \textbf{\DATASET}, a benchmark reflecting the distinct goal of educational videos: teaching new knowledge. It comprises professionally produced, discipline-specific Manim tutorials across 13 domains (\eg~topology, physics).
Evaluation covers three complementary dimensions: (i) VLM-as-a-Judge aesthetic and structural quality;
(ii) code efficiency, measuring generation time and token consumption; 
and (iii) \textbf{\quiz}, a novel end-to-end knowledge-transfer metric that enforces unlearning of the target concept in a VLM, and then measures how effectively the generated video restores it.
This multi-dimensional protocol directly probes educational efficacy and grounds a code-centric paradigm for video generation.
Our results reveal clear trends: pixel-based models struggle with fine details and coherence, while direct code-centric generation improves \quiz by 30\%. Our full Planner–Coder–Critic pipeline further delivers a stable 40\% gain. In human studies on \quiz scores, agentically generated videos even outperform professional human-made tutorials, underscoring \textit{the effectiveness of code-centric, agentic generation}.

Our contributions are summarized as follows:
\begin{itemize}[leftmargin=*]
    \vspace{-0.2cm}
    \item \textbf{A New Paradigm for Video Generation.} We introduce a new code-centric paradigm for educational video generation, positioning executable code as the unifying medium for temporal sequencing and spatial organization.
    \item \textbf{Effective Designs for Visual Animation Agent.} 
    {We highlight a modular agent design with three key components: (i) Planner expands an external database for reference, enabling parallel yet consistent storyboard; (ii) Coder ensures compilable code via automatic debugging and scope-guided repair; (iii) Critic refines spatial layout and clarity using visual anchor prompts.}
    \item \textbf{New Benchmark with Well-designed Evaluation.} We present {\DATASET}, the first benchmark for code-centric educational video generation with multi-dimensional evaluation of efficiency, aesthetics, and end-to-end knowledge transfer.
\end{itemize}

\section{Related Work}



\subsection{Video Generation}
Early text-to-video generation methods \textbf{(\romannumeral1)} extend diffusion models into the temporal domain via space–time UNets and latent 3D VAEs~\citep{weng2024genrec, ho2022video}, achieving strong perceptual fidelity and longer durations~\citep{yang2024cogvideox, li2024survey, xing2024make}. However, their reliance on \textit{pixel-space} synthesis limits controllability, making precise layout and symbolic alignment—both critical for educational videos—difficult to realize. Autoregressive and progressive schemes~\citep{li2024arlon, gu2025long, wang2024loong, xie2025progressive} have improved long-form generation~\citep{lu2024freelong, zhou2024storydiffusion}, yet still struggle with board-like composition and stepwise exposition required in educational contexts~\citep{li2024survey, liu2024sora}.
\textbf{(\romannumeral2)} Recent advances in \textbf{multi-agent collaboration} show that decomposing tasks, coordinating tool use, and enabling iterative self-improvement can substantially enhance reasoning and generation~\citep{yuan2024mora, hu2024storyagent, xie2024dreamfactory, shen2024data}. 
While multi-agent frameworks have proven effective in domains such as web interaction, their application to video generation remains unexplored~\citep{ku2025theoremexplainagent, wu2024autogen}.
\textbf{(\romannumeral3)} Building on this paradigm, we propose a \textbf{code–centric animation framework} for educational video synthesis. By elevating executable code as the generative substrate, our approach achieves symbolic layout, temporally structured exposition, and deterministic reproducibility—capabilities unattainable with pixel-level diffusion.

\subsection{Coding Agents}

Recent advances in LLM-based tool use have shown that agents can autonomously call APIs, retrieve specifications, and verify outputs, enabling neuro-symbolic modularity and robust task decomposition~\citep{yao2023react, wang2025toward}. By integrating code execution and tool invocation, representative methods extend language models beyond \textbf{text-only} reasoning, supporting complex workflows and project-level code generation~\citep{patil2024gorilla, liu2025projecteval, gupta2024codenav}. Such developments demonstrate the potential of LLM agents to coordinate external retrieval, maintain memory across parallel processes, and incorporate feedback loops for iterative refinement~\citep{li2025review, xu2025llm, zhang2024vipact}.
In parallel, research at the intersection of coding and visual reasoning shows that generating and executing programs can yield structured perception and controllable rendering~\citep{pang2025paper2poster, zhu2025uniapo, lin2025showui}. \textbf{Visual programming} and visual-to-code approaches leverage program synthesis for compositional reasoning and spatial arrangement, with benchmarks translating images or text into executable code for charts, plots, and graphical interfaces~\citep{wu2024plot2code, zhao2025chartcoder, wei2025words, yen2025code}. While these works bridge symbolic and visual domains, they largely focus on \textit{static} figures or localized visual tasks~\citep{xing2025chartcode, wen2024program, ye2025mographgpt, jain2025manimator}.
We advance this line by integrating code generation and visual synthesis for \textit{dynamic} educational \textbf{video creation}. 

\section{\DATASET~Benchmark}

\begin{figure}[htb]
  \centering
  \includegraphics[width=\linewidth]{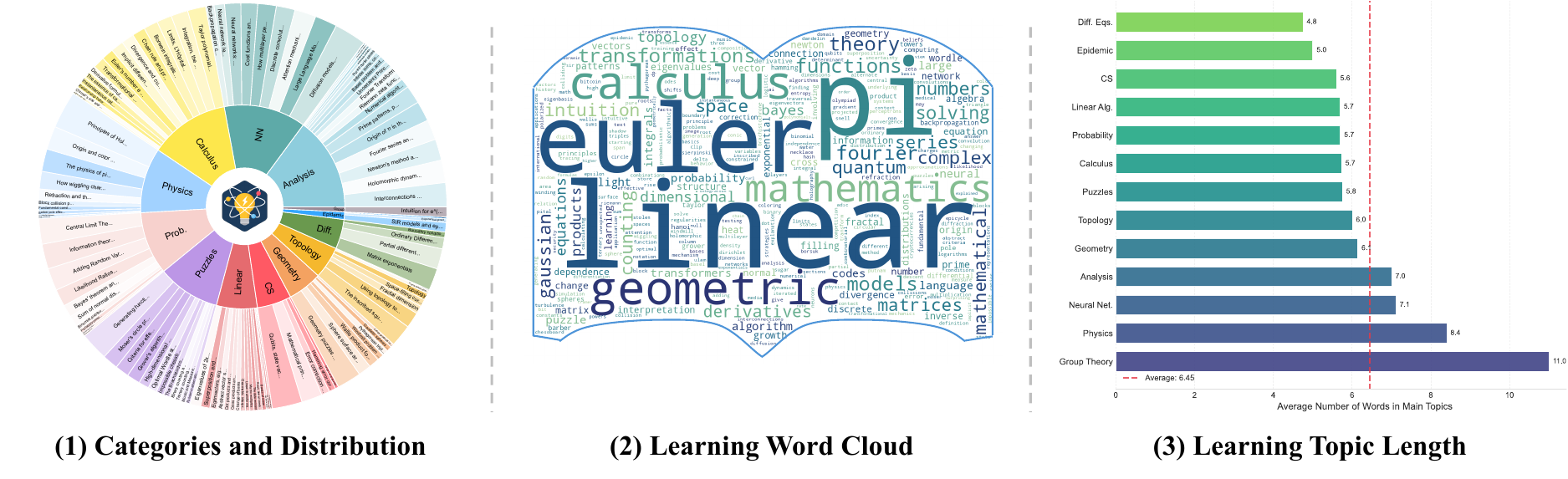}
  \caption{ \textbf{\DATASET overview}. (1) Left: distribution of 13 subject categories with exemplar learning topics; ring width encodes video duration. (2) Middle: learning topic word cloud highlighting core concepts. (3) Right: average learning topic length per category.}
  \label{dataset}
\end{figure}

\subsection{Task Formulation} 
The task of code-centric educational video generation maps a learning query to executable \textit{Manim}~\citep{manim2025code} code whose rendering yields a tutorial video. The challenge lies in multi-step reasoning, precise temporal sequencing, and spatial coherence, where even minor syntax errors can nullify execution. We adopt \textit{Manim} for its fine-grained spatiotemporal control, symbolic expressivity, and demonstrated effectiveness in expert-produced instructional videos.

\subsection{Data Curation and Statistics}

We construct {\DATASET}, a benchmark for code-driven educational video generation, under two criteria: (i) \textit{educational relevance}—each learning topic is an established concept worth teaching; and (ii) \textit{executable grounding}—each concept aligns with a high-quality Manim reference, ensuring practical realizability.  
We source from the complete 3Blue1Brown (3B1B) YouTube corpus, known for its instructional impact and expert Manim craftsmanship. After filtering out non-instructional items (e.g., Q\&A), we curate 117 long-form videos spanning 13 subject areas, including \textit{calculus}, \textit{geometry}, \textit{probability}, and \textit{neural networks}. To enrich supervision, we segment videos using author-provided timestamps into 339 semantically coherent sub-clips, yielding 456 units in total. An LLM then extracts concise learning topics (avg.~6.3 words) from titles, descriptions, and metadata, producing a clean mapping from videos to educationally grounded units (details in \S\ref{datadetails}).  
On average, a full-length video lasts 1014 seconds ($\sim$16.9 minutes), while a segmented clip spans 201 seconds ($\sim$3.35 minutes), thus balancing long-horizon reasoning with fine-grained supervision. Figure~\ref{dataset} visualizes topical diversity with a hierarchical donut plot: the inner ring denotes 13 categories, while the outer ring shows individual topics with arc width proportional to cumulative duration. This structure highlights both the breadth of coverage and the temporal richness of {\DATASET}, establishing it as a challenging and representative benchmark for educational video generation.

\subsection{Evaluation Metrics}  
Unlike conventional video generation, educational videos are valued less for visual fidelity than for how effectively they convey knowledge. This makes standard synthesis metrics inadequate. We therefore design a three-pronged evaluation across \textbf{aesthetics}, \textbf{knowledge convey}, and \textbf{efficiency}:

\vspace{-0.3cm}
\paragraph{VLM-as-Judges.} 
Since human judgments of video quality are inherently subjective, we adopt a {VLM-as-judges} protocol (\hyperref[p_aes]{$\mathcal{P}_{\rm aesth}$}) to approximate user perception across five axes:  
\textit{(\romannumeral1)~Element Layout (EL)} — clarity and spatial arrangement of visual components.  
\textit{(\romannumeral2)~Attractiveness (AT)} — overall engagement and ability to capture learners' attention.  
\textit{(\romannumeral3)~Logic Flow (LF)} — coherence in temporal presentation of concepts.  
\textit{(\romannumeral4)~Visual Consistency (VC)} — stylistic stability across frames and sections.  
\textit{(\romannumeral5)~Accuracy \& Depth (AD)} — correctness and richness of the presented knowledge.  
Each dimension is rated on a 100-point scale.

\vspace{-0.2cm}
\begin{figure}[htb]
  \centering
  \includegraphics[width=1.0\linewidth]{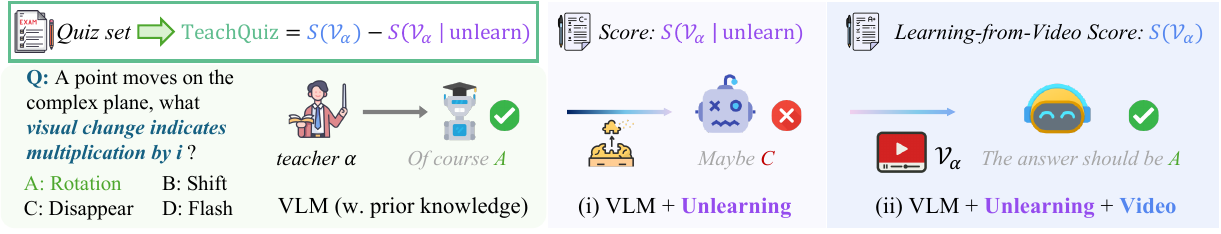}
  \caption{{\quiz}: score gap between \textit{Learning-from-Video} and \textit{Unlearning} stages.}
  \label{fig_quiz}
\end{figure}
\vspace{-0.3cm}
\paragraph{TeachQuiz.}
The goal of educational video generation is not merely visual plausibility, but effective knowledge transfer. To evaluate this, we introduce {TeachQuiz}, a two-stage protocol grounded in a quiz set $\mathcal{Q}(\mathcal{K})={(q_i, y_i)}_{i=1}^N$ for a given concept $\mathcal{K}$, and $Y$ denotes ground-truth answers. We consider multiple teachers $\alpha, \beta$, each producing a video $\mathcal{V}_\alpha, \mathcal{V}_\beta$. A student model $\phi$ is tasked with watching the video and answering questions:
\begin{equation}
    S(\mathcal{V}_\alpha)=\mathbf{1} [\phi \left(Q, \mathcal{V}_\alpha\right)=Y]
\end{equation}
If $S(\mathcal{V}_\alpha) > S(\mathcal{V}_\beta)$, then teacher $\alpha$ is the stronger instructor.

However, a key challenge is that \textit{many quiz items are already be learn by top-performing VLMs (\ie~\textbf{answer correctly without watching the video})}. Thus, absolute accuracy alone does not measure teaching quality. Instead, a good educational video should improve knowledge acquisition relative to a controlled baseline. We enforce this through two steps:
\textbf{(i) Unlearning.} Apply \hyperref[p_unlearn]{$\mathcal{P}_{\rm unlearn}$} to block prior access to $\mathcal{K}$, yielding a knowledge-removed baseline.
\textbf{(ii) Learning-from-Video.} Expose the model to $\mathcal{V}$ under \hyperref[p_learn]{$\mathcal{P}_{\text{learn}}$}, testing whether the video itself enables recovery of the knowledge.
The final \textit{TeachQuiz} score measures relative improvement:
\begin{equation}
\widetilde{S}(\mathcal{V}_\alpha)=S(\mathcal{V}_\alpha) -S(\mathcal{V}_\alpha|\text{unlearn})
\end{equation}
which isolates the contribution of the video by subtracting the unlearned baseline. Higher $\widetilde{S}$ indicates stronger knowledge transfer induced by the generated video.

\vspace{-0.3cm}
\paragraph{Token Cost and Generation Time.}
Beyond output quality, an equally important dimension is how economically a model can generate effective videos. We measure efficiency by \textit{average code generation time} and \textit{token usage per video}, reflecting scalability and feasibility in large-scale or interactive educational settings where latency and resource costs are critical.


\section{Method: \MODEL}
\label{sec_approach}
\begin{figure}[!t]
  \centering
  \includegraphics[width=1.0\linewidth]{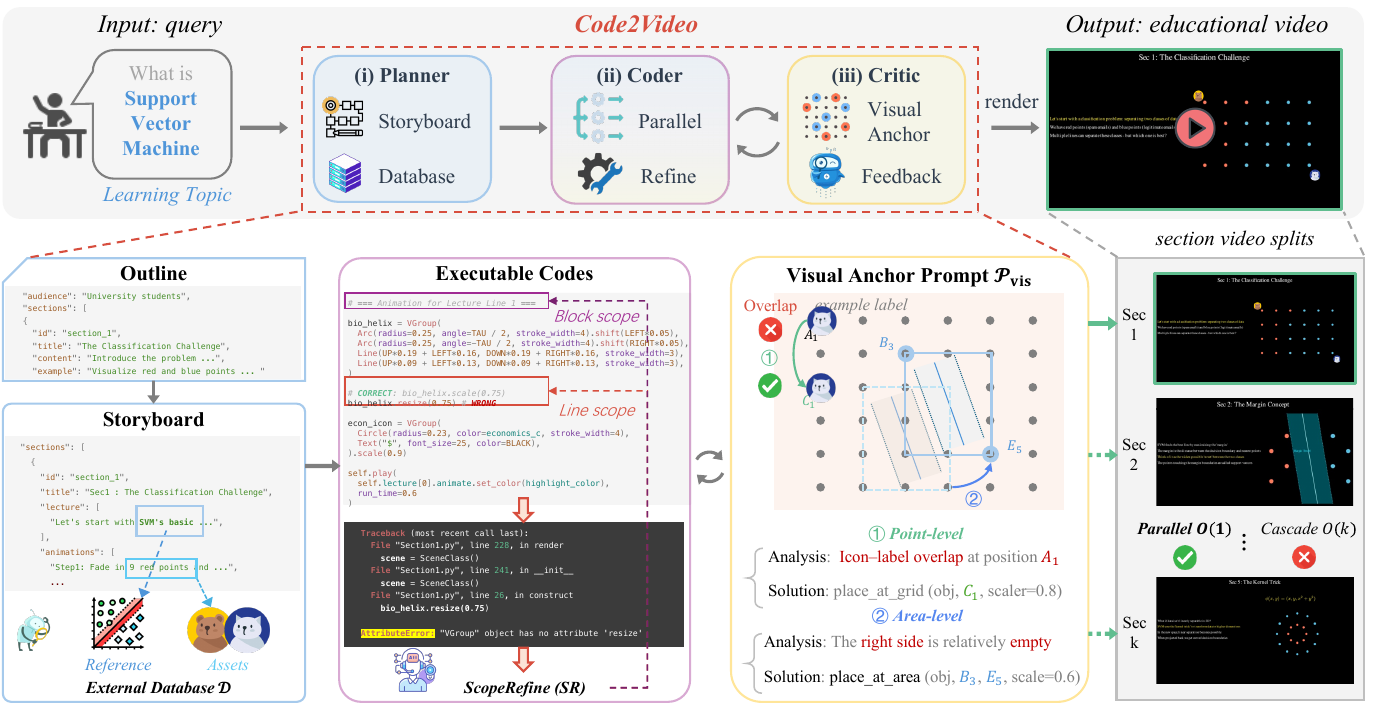}
  \caption{\textbf{Illustration of \MODEL.} Given a user inquiry, \our aims to render an educational video via Manim code writing:
  \textbf{(i) the Planner} converts a learning topic into a storyboard and retrieves visual assets; \textbf{(ii) the Coder} performs parallel code synthesis with scope-guided refinement to ensure efficiency and temporal consistency; \textbf{(iii) the Critic} uses visual anchor prompts to iteratively adjust spatial layout and clarity, yielding reproducible, educationally structured videos.
  }
  \label{fig_approach}
\end{figure}

\textbf{Overview.}  
As illustrated in Fig.~\ref{fig_approach}, given a topic query $\mathcal{Q}$, \MODEL~output a video $\mathcal{V}$, which is consists of three stages: 
\textbf{(i) Planner} structures topics into storyboards with reference assets, \textbf{(ii) Coder} translates each section into executable Manim code using parallel synthesis and an effective debugging, and \textbf{(iii) Critic} refines rendered videos through a novel visual prompt and VideoLLM feedback to ensure spatial coherence and educational clarity.

\subsection{Planner: query to storyboard}

Generating coherent educational videos requires careful organization of temporal structure. We design the Planner to decompose a topic $\mathcal{Q}$ into two stages: {outline generation} for high-level ordering, and {storyboard construction} for stepwise realization. This preserves logical flow while capturing cross-section dependencies.

\textbf{(i) Outline Generation.} Given a topic $\mathcal{Q}$, the Planner produces an outline $\mathcal{O} = {o_1, \dots, o_n}$, where each $o_i$ contains a unique identifier, section title, content summary, and illustrative examples. Crucially, the Planner also considers the intended audience (\eg~trigonometric functions for middle school, Fourier’s law for undergraduates), ensuring level-appropriate structure.  Formally, $\mathcal{O} \leftarrow \hyperref[p_outline]{\mathcal{P}_{\rm outline}}(\mathcal{Q})$, where $\mathcal{O}=\{o_1, \cdots, o_n\}$ and each $o_i$ encodes the section-level metadata and educational intent. By explicitly specifying audience and structure, the outline establishes the temporal skeleton for the subsequent video, guiding both pacing and sequencing.

\textbf{(ii) Storyboard Construction.} The second stage converts the outline $o$ into a detailed storyboard $s$. Each section in $s$ includes title, lecture lines, and corresponding animations, with $s_i \leftarrow \hyperref[p_storyboard]{\mathcal{P}_{\rm storyboard}}(o_i)$. The storyboard specifies the temporal sequence of lecture lines and paired animations, bridging high-level planning with concrete visual content.

\textbf{External Database.} 
To enhance factual accuracy and visual fidelity, the Planner integrates an external database $\mathcal{D}$. It includes \textit{(a) reference images} aligned with the topic to anchor complex concepts and reduce hallucination, and \textit{(b) visual assets} (\eg~icons, logos) that are difficult to generate from scratch. These assets $\mathcal{A}$ are automatically identified via a prompt \hyperref[p_assets]{$\mathcal{P}_{\rm asset}$} analyzing the storyboard, $a_i \leftarrow \mathcal{P}_{\rm asset}(s_i)$, and stored in a persistent cache $\mathcal{D}_{\rm asset}$. 
Caching enables reuse across sections, preventing redundant generation and ensuring visual consistency.
{Please refer to \S~\ref{appendix_assets} for more details and examples about $\mathcal{D}$.}

\subsection{Coder: Storyboards to Executable Code}

The Coder $\mathcal{G}$ translates each section of the storyboard $s$ and the cached assets $a$ into executable Manim code $C=\{c_1,\dots,c_n\}$, where each $c_i$ corresponds to a storyboard $s_i$. 

\textbf{(i) Parallel Code Generation.}  
A central bottleneck in full-code synthesis is generation time, as end-to-end Manim code production for a single educational video—including generation, debugging, and rendering—can exceed two hours. 
To address this bottleneck, we parallelize the pipeline by decoupling serial steps—code generation, debugging, and refinement—so that each section is synthesized and fixed independently.
Each section is conditioned on its storyboard and shared assets $\mathcal{A}$: $c_i = \hyperref[p_coder]{\mathcal{P}_{\rm coder}}(s_i, \mathcal{A})$. 
Notably, asset sharing across sections ensures temporal consistency while retaining the efficiency benefits of parallelization.

\textbf{(ii) Effective Debugging.} 
Even strong LLMs rarely produce fully executable code in one pass. Naïve strategies that concatenate all code with the full error log are costly in both time and tokens. We propose \textbf{ScopeRefine (SR)}, a hierarchical, scope-guided repair strategy, as illustrated in Fig.\ref{fig_approach} middle bottom:
\textit{(a) Line scope:} isolate the error line plus immediate context, $\mathcal{S}_1={\rm line}\pm1$, attempt up to $K_1$ local fixes.  
\textit{(b) Block scope:} if unresolved, expand to the lecture-line block $\mathcal{S}_2=\mathcal{B}_{i,j}$ with up to $K_2$ repair attempts.  
\textit{(c) Global scope:} as a last resort, regenerate the full section $c_i$ from $s_i$.  
This progressive \textit{``Go-to style'' repair} minimizes token usage and latency while ensuring high reliability, effectively bridging parallel generation with robust debugging.


\subsection{Critic: Effective Visual Refinement}


\begin{wrapfigure}[15]{r}{0.63\textwidth}
    \vspace{-12pt}
    \centering
    \includegraphics[width=\linewidth]{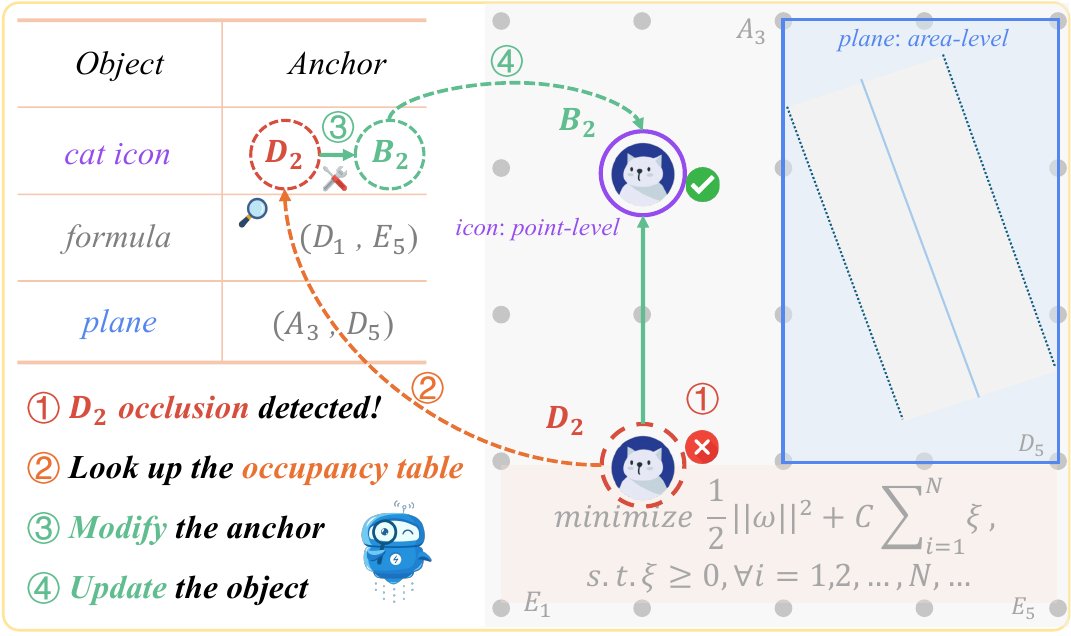}
    \vspace{-17pt}
    \caption{Illustration of \textit{visual anchor prompt ({$\mathcal{P}_{\rm vis}$})}.}
    \label{fig_anchor}
\end{wrapfigure}

{Even after debugging ensures executability, the generated code may still yield unsatisfactory visual outcomes. LLMs and VLMs often fail to provide actionable feedback due to \textbf{limited spatial awareness}~\citep{cheng2024spatialrgpt, zha2025enable}. In practice, models can identify issues (\eg~“the cat icon is misplaced”) but struggle to provide actionable corrections. They often fail to indicate the direction or distance needed to adjust the element, which makes text-only refinement inadequate.


\textbf{(i) Visual Anchor Prompt (\hyperref[p_vis]{$\mathcal{P}_{\rm vis}$})}. We introduce $\mathcal{P}_{\rm vis}$, a textual prompt that discretizes the 2D canvas into a $6 \times 6$ grid of predefined anchor points. Each grid cell is mapped to fixed Manim coordinates, allowing LLM-specified locations to be directly converted into executable code. Placement follows two granularities, as illustrated in Figure~\ref{fig_anchor}: \textcolor{googlepurple}{{\textit{(a) point-level}}}, where small elements (\eg~symbols, short labels) occupy a single anchor; and \textcolor{googleblue}{\textit{(b) region-level}}, where larger elements are assigned to a bounding box spanning multiple anchors. 
 This discretization transforms the placement task from a \textit{continuous positioning problem} into a \textit{discrete anchoring problem}, serving as a visual debugging \textit{``go-to''}, which substantially reduces the difficulty for LLMs to produce valid layouts.

\textbf{(ii) VideoLLM for Code Feedback.} 
To detect violations and refine placement, the Critic inspects the rendered video $\mathcal{V}_i$ alongside its section code $c_i$. During parallel code generation, we maintain an \textit{occupancy table} that records each element’s assigned anchors (point or region), scaling factor, and corresponding code lines. This design serves two purposes: (a) it makes all assets indexable, allowing the Critic to quickly trace a visual issue back to its source code; and (b) it reveals available anchors, enabling conflict-free reallocation.
With this structured view, the Critic efficiently detects three common issues: overlapping elements within a cell, lecture lines occluded by animations, and large unused regions creating visual imbalance. These findings are incorporated into a refinement prompt \hyperref[p_refine]{$\mathcal{P}_{\rm refine}$}, yielding optimized code:
\(\tilde{c}_i = \mathcal{P}_{\rm refine}(c_i, \mathcal{V}_i)\)
and final video \(\widetilde{\mathcal{V}} = \text{Render}(\{\tilde{c}_i\}_{i=1}^n)\). By combining anchor-based guidance, indexable, occupancy-aware adjustment, and multimodal feedback, the Critic overcomes the limitations of text-only debugging.

\section{Experiment}

\subsection{Implementation Details}
\textbf{Baselines.} We compare four types of approaches: 
$\diamond$ \textit{{Human-crafted}}, expert-designed Manim videos as an upper bound; 
$\diamond$ \textit{Pixel-based Diffusion}, text-to-video models:~\textit{OpenSora-v2}~\citep{peng2025open}, \textit{Wan2.2-T2V-A14B}~\citep{wan2025wan}, and \textit{Veo3}~\citep{veo3}; 
$\diamond$ \textit{{CodeLLM Generation}}, where an LLM directly generates Manim code from a learning topic;
$\diamond$ \textit{{Agentic Generation} (\textbf{ours})}, a Planner–Coder–Critic pipeline.
We evaluate across diverse models: \textit{Claude Opus 4.1}~\citep{anthropic2025claudeopus}, \textit{GPT-4o}, \textit{GPT-o4 mini}, \textit{GPT-4.1}, \textit{GPT-5}~\citep{openaigpt}, \textit{Gemini-2.5 Pro}~\citep{imran2024google}, with \textit{Gemini-2.5 Pro} serving as Critic for refinement.
\textbf{Evaluation.} Aesthetics are judged by \textit{Gemini-2.5 Pro} (VLM-as-a-Judge), and quantify knowledge transfer with \quiz.
\textbf{Resources.} Reference images are retrieved from Google Images, and visual assets from Iconfinder\footnote{https://www.iconfinder.com}. 
All prompts are documented in \S~\ref{prompts}.

\subsection{Main Results}

\setlength{\tabcolsep}{4.5pt}
\begin{table}[htb]
\caption{Results across Efficiency, Aesthetics, and \quiz (Quiz). Efficiency: Time (\textbf{avg minutes} per topic) and Token (avg \textbf{token consumption} per topic). Aesthetics: Element Layout (EL), Attractiveness (AT), Logic Flow (LF), Visual Consistency (VC), Accuracy \& Depth (AD).}
\label{tbl_results}
\centering
\small
\begin{tabular}{lcccccccll}
\toprule
\multirow{2}{*}{Method} & \multicolumn{2}{c}{Efficiency~($\downarrow$)} & \multicolumn{6}{c}{Aesthetics ($\uparrow$)} & \multirow{2}{*}{Quiz ($\uparrow$)} \\
\cmidrule(r){2-3}
\cmidrule(r){4-9}
           & Time & Token (K) & EL & AT & LF & VC & AD & \textbf{Avg} &  \\
\midrule
\rowcolor{gray!10} 
Human-made 3B1B& \multicolumn{1}{c}{--}  & \multicolumn{1}{c}{--} & 98.3 & 100 & 100 & 100 & 100  & 99.7 & 97.1 \\
\midrule
\multicolumn{10}{c}{{\textcolor[RGB]{105, 105, 105}{\textit{Pixel-based Diffusion}}}}  \\
OpenSora-v2 & 27.6 & \multicolumn{1}{c}{--}  & 0.0 & 5.0 & 0.0 & 0.0 & 13.3 & 3.7 & 0.0 \\
Wan2.2-T2V-A14B & 17.4 &  \multicolumn{1}{c}{--} & 0.0 & 10.0 & 0.0 & 0.0 & 20.0 & 6.0 & 0.0 \\
Veo3 & 2.3 & -- & 0.0 & 15.0 & 0.0 & 5.0 & 25.0 & 9.0 & 2.5 \\
\midrule
\multicolumn{10}{c}{{\textcolor[RGB]{105, 105, 105}{\textit{Code LLM}}}}  \\
GPT-5 & 1.8 & 1.1 & 27.0 & 28.0 & 28.0 & 54.5  &26.0   & 32.7 & 36.5 \\
GPT-4.1 & 2.1 & 1.2 & 30.5 & 34.5 & 39.0 & 42.0 & 24.8 & 34.2 & 37.0 \\
Claude Opus 4.1 & 2.8 & 2.3 & 47.5 & 40.0 & 26.5 & 56.6 & 18.4 & 37.8  & 40.0 \\
\midrule
\multicolumn{10}{c}{{\textcolor[RGB]{16,185,129}{\textit{\MODEL~Agent (Ours)}}} } \\
\MODEL\textsubscript{{\textbf{Gemini-2.5 Pro}}} &  15.5& 41.8 &70.3&60.3&44.3&37.6&74.7& 57.4 & 72.0 \\
\MODEL\textsubscript{{\textbf{GPT-4o}}} & 14.1& 32.7 &70.3&58.3&54.6&48.5&68.3& 60.0 & 44.0 \\
\MODEL\textsubscript{{\textbf{GPT-o4 mini}}} & 16.8&49.2 & 77.0& 52.8 &73.0 &57.2 &79.0 & 67.8 & 48.5 \\
\MODEL\textsubscript{{\textbf{GPT-5}}} &  8.8 &19.3 & 75.5 & 60.5 & 81.8 & 63.6 & 79.7 & 72.2~\textsubscript{\textcolor{vscodegreen}{+39.5}} & 80.0~\textsubscript{\textcolor{vscodegreen}{+43.5}} \\
\MODEL\textsubscript{{\textbf{GPT-4.1}}} & 15.4 & 30.8 & 82.8 & 65.6 & 95.0 & 68.0 & 83.7 & 79.0~\textsubscript{\textcolor{vscodegreen}{+44.8}} & 82.0~\textsubscript{\textcolor{vscodegreen}{+45.0}} \\
\MODEL\textsubscript{{\textbf{Claude Opus 4.1}}} & 13.8  & 43.1 & 90.6 & 79.7 & 93.3 & 84.2 & 91.9 & \textbf{87.9}~\textsubscript{\textcolor{vscodegreen}{+50.1}} & \textbf{86.0}~\textsubscript{\textcolor{vscodegreen}{+46.0}} \\
\bottomrule
\end{tabular}%
\end{table}

Table~\ref{tbl_results} compares {\MODEL} with human-crafted videos, pixel-based models, and code LLM baselines, evaluated on Efficiency, Aesthetics (AES), and knowledge transfer (\quiz).
Our analysis yields several insights:
\textbf{(\romannumeral1) Pixel-based models underperform.} They obtain the lowest scores on both AES and \quiz, particularly struggling with LF due to weak control over text grounding, animation timing, and cross-frame coherence.
\textbf{(\romannumeral2) Direct code-centric generation delivers clear improvements}. Rendering videos from LLM-produced Manim code outperforms pixel-based models, underscoring code as an effective medium for controllable and coherent educational video generation.
\textbf{(\romannumeral3) Our agentic framework delivers stable and consistent improvements.} Across different backbone LLMs, {\MODEL} achieves significant performance boosts. For instance, with Claude Opus 4.1, AES improves by 50\% and \quiz by 46\%. These gains arise from distinct components: visual anchor points drive improvements in element layout, while the Planner enhances LF and AD. However, limitations remain in AT and VC, pointing to opportunities for refinement.
\textbf{(\romannumeral4) Human-made videos remain strong.} Although {\MODEL} narrows the gap, professional videos still lead in storytelling, nuanced sequencing, and explanatory depth. This highlights the next frontier: advancing agentic pipelines toward \textbf{\textit{professional-quality long educational videos}}.

\paragraph{Qualitative Analyses.} 
Figure~\ref{fig_compare_main} illustrates that our code-driven pipeline produces videos with clear text and formulas, stable layouts without occlusions, and stepwise alignment with lecture lines. In contrast, the pixel-based model (Veo3) often generates blurry or corrupted text, inconsistent styles, and drifting visuals, weakening semantic grounding. Overall, code-driven synthesis ensures better spatial stability and clearer knowledge presentation. Additional cases are provided in \S~\ref{quality}.

\begin{figure}[t!]
  \centering
  \includegraphics[width=1.0\linewidth]{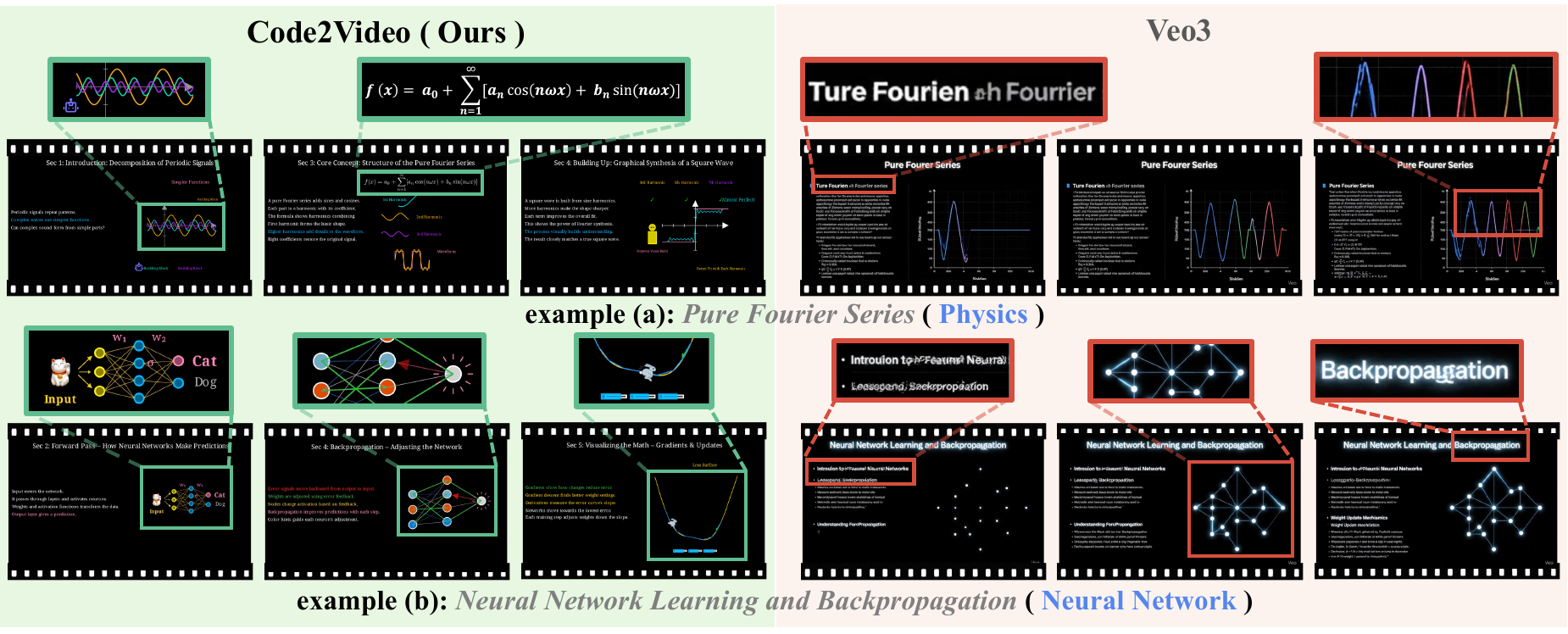}
  \caption{{Qualitative comparison} between \textit{\MODEL} and \textit{Veo3}. Our approach generates videos with coherent logic flow, consistent semantics, and interpretable layouts.}
  \vspace{-8pt} 
  \label{fig_compare_main}
\end{figure}

\subsection{Ablation Studies}


\paragraph{Effects by Individual Components.} 
Table~\ref{tbl_sub1} highlights several observations.
First, \quiz is more sensitive than Aesthetics, revealing \textit{knowledge-transfer gaps even when videos still look visually acceptable}.
Second, the Planner is essential: removing it collapses both metrics ($\approx$~41 points), underscoring that high-level lecture planning and temporal sequencing are the backbone of effective teaching videos.
Third, other modules provide complementary gains: the External Database improves conceptual grounding, Visual Anchors stabilize layouts, and the Critic ensures refinement—each modest alone, but jointly essential for robustness.
These results \textbf{highlight that structured visual guidance and iterative refinement are crucial} for producing visually clear videos that effectively convey knowledge.

\begin{table*}[htb]
\setlength{\tabcolsep}{3.pt}
\begin{floatrow}
\capbtabbox{
\small
\begin{tabular}{lcc}
\toprule
 Method  & Aesthetics & Quiz  \\
\midrule
\rowcolor{lightgreen}
\MODEL\textsubscript{{\textbf{Chat-4.1}}} ($\diamond$) & \textbf{79.0} & \textbf{82.0} \\
$\diamond$~{w/o Planner}  & 38.1~\red{$_{-\mathbf{40.9}}$} & 40.5~\red{$_{-\mathbf{41.5}}$} \\
$\diamond$~{w/o External Database}  & 68.1~\red{$_{-10.9}$} & 52.0~\red{$_{-30.0}$}  \\
$\diamond$~{w/o Visual Anchor}  & 69.2~\red{$_{-9.8}$} & 55.2~\red{$_{-26.8}$}\\
$\diamond$~{w/o {Critic}} & 72.5~\red{$_{-6.5}$} & 60.7~\red{\textsubscript{${-21.3}$}}  \\

\bottomrule
\end{tabular}
}
{
 \caption{Effect of different components on quality: \quiz~/ Aesthetics avg. score.}
 \label{tbl_sub1}
}
\capbtabbox{
\small
\begin{tabular}{lcc}
\toprule
Method & Time (m) & Token (K)\\
\midrule
\rowcolor{lightgreen}
\MODEL\textsubscript{{\textbf{Chat-4.1}}}~($\diamond$)  & \textbf{15.4} &  \textbf{30.8}  \\
$\diamond$~{w/o parallel} & 86.6~\red{$_{5.6\times}$} & 30.8 \\
$\diamond$~{w/o SR $\rightarrow$ w. Retry} & 42.9~\red{$_{2.8\times}$} & 49.8~\red{$_{1.6\times}$} \\
$\diamond$~{w/o SR $\rightarrow$ w. Debug} & 39.2~\red{$_{2.5\times}$} & 42.1~\red{$_{1.4\times}$} \\
$\diamond$~{w/o parallel \& SR} & 149.8~\red{$_{\mathbf{9.7\times}}$}  & 52.6~\red{\textsubscript{$\mathbf{1.7\times}$}} \\
\bottomrule
\end{tabular}
}
{
 \caption{Effect of efficiency components: runtime avg. time / token consumption.}
 \label{tbl_sub2}
}
\end{floatrow}
\end{table*}

\vspace{-0.3cm}

\paragraph{Efficiency Components.} Table~\ref{tbl_sub2} evaluates efficiency-oriented modules. Removing parallel execution greatly increases latency (\(15.4 \to 86.6\) minutes). 
Without ScopeRefine (SR), we test two alternative debugging methods: (i) \textit{Retry}, which regenerates the section upon any error; (ii) \textit{Full-code Debug}, which feeds the entire code and error log to the LLM to regenerate the section. In both cases, error correction is costly, highlighting the importance of SR’s localized, scope-aware repair.
Removing both mechanisms produces prohibitive overheads. These results underscore that parallel synthesis and scope-aware repair are essential for scalable, code-centric video generation.

\setlength{\tabcolsep}{3.8pt}
\begin{table}[htb]
\caption{\textbf{Human study} on Aesthetics, TeachQuiz (Quiz), Completion Willingness (CW), and Average Ranking (AR). Results align with VLM-based trends but show sharper score contrast, lower tolerance for layout errors, and reduced engagement in longer-duration videos.}
\label{tbl_human}
\centering
\small
\begin{tabular}{lcccccccccc}
\toprule
\multirow{2}{*}{Method} & \multirow{2}{*}{Duration} &  \multicolumn{6}{c}{Aesthetics ($\uparrow$)} & \multirow{2}{*}{Quiz ($\uparrow$)} & \multirow{2}{*}{CW ($\uparrow$)} & \multirow{2}{*}{AR ($\downarrow$)} \\
\cmidrule(r){3-8}
         & & EL & AT & LF & VC & AD & Avg &  \\
\midrule
\textcolor{gray}{Human-made 3B1B} & 16.9 min & \cellcolor{lightgreen}{$98.9$} & 97.2 & 91.3 & 98.0 & 97.0 & 96.5  & \cellcolor{lightred}{$78.8$} & \cellcolor{lightred}{$36.2$} & \cellcolor{lightgreen}{$1.2$} \\
Pixel-based~\textsubscript{\textbf{Veo3}} & 8.0 s & 12.6 & 4.4 & 1.1 & 24.4 & 1.1 & 8.5 & 8.0 & 46.8 & \cellcolor{lightred}{$5.0$} \\
Code LLM~\textsubscript{\textbf{Claude Opus 4.1}} & 0.9 min & 16.1 & 41.1 & 55.6 & 71.1 & 72.2 & 51.2 & 56.6 & 15.0 & 3.9 \\
\MODEL\textsubscript{{\textbf{Gemini-2.5 Pro}}}  & 1.6 min & \cellcolor{lightred}{$26.7$} & 68.3 & 78.1 & 90.2 & 81.0 & 68.9  & \cellcolor{lightred}{$65.3$} & 47.4 & 3.1 \\

\MODEL\textsubscript{{\textbf{Claude Opus 4.1}}} & 2.0 min & \cellcolor{lightred}{$60.2$} & 89.3 & 84.6 & 92.0 & 83.1 & 81.8 & \cellcolor{lightgreen}{$\mathbf{80.3}$} & \cellcolor{lightgreen}{$\mathbf{64.0}$} & \cellcolor{lightgreen}{$1.8$} \\
\bottomrule
\end{tabular}%
\end{table}
\vspace{-0.3cm}

\paragraph{Human Study Evaluation.} 
We conduct a five-group user study (6 middle school, 2 undergraduate volunteers per group), where each participant watches one video type and answers 5 quiz questions for 20 learning topics. We measure Completion Willingness (\textbf{CW}, proportion finishing the video before answering, max score is 100) and Average Ranking (\textbf{AR}, mean preference across video types, 1 is the best).
Table~\ref{tbl_human} reveals four patterns: \textbf{(i) Clearer separation.} Human ratings follow the same overall trends as VLM-based scores but with stronger contrast: high-quality videos are rated in the upper range ($>90$), while low-quality videos cluster near the lower bound ($<10$). \textbf{(ii) Sensitivity to layout errors.} Participants give lower layout scores (EL) to videos from \MODEL, as humans are highly sensitive to even brief occlusions, whereas VideoLLMs often miss such frame-level issues. \textbf{(iii) Attention span limits.} Human attention is inherently limited: to perform well on the quiz, participants must follow the full flow of knowledge details in the video. This requires not only \textit{strong logical coherence} and \textit{engaging presentation} but also a \textit{reasonable duration} that allows sustained high attention for effective knowledge absorption.
\textbf{(iv) Strong consistency.} Aesthetics and \quiz scores are strongly correlated($r=0.971$, $p=0.0059$): visually appealing videos keep students engaged, leading to higher learning outcomes.
Overall, the human study underscores that both structural clarity and visual appeal are decisive levers for learning efficacy, complementing the automated metrics.
{\textit{Future work requires agent designs that explicitly account for \textbf{human attention and patience}, ensuring videos maintain \textbf{fine-grained details} while \textbf{minimizing perceptual fatigue}.}} 

\section{Conclusion}


We have introduced a novel, code-centric paradigm for educational video generation, establishing executable code as the unifying medium for both temporal sequencing and spatial organization. Building on this paradigm, our tri-agent architecture \textit{\MODEL} enables controllable and interpretable generation with multimodal feedback. 
To systematically evaluate this paradigm, we introduce \textit{\DATASET}, targeting efficiency, aesthetics, and knowledge transfer. 
Together, our paradigm, architecture, and benchmark chart a clear path for future research on leveraging code as a medium for high-quality, structured, and interpretable educational content generation.
Future work includes broadening the video scope and developing more lightweight, scalable agent frameworks.


\bibliography{A-reference}
\bibliographystyle{iclr2026_conference}

\newpage

\appendix

\section{Supplementary Material}

\subsection{Additional Implementation Details and Experiments}

\subsubsection{Unlearning Details and \quiz}
\label{unlearn}

To probe whether generated tutorial videos genuinely transfer knowledge, we integrate a selective unlearning–relearning protocol into the \quiz evaluation.

\textbf{Model choice.} We adopt \textit{Gemini-2.5 Pro}~\citep{imran2024google}, one of the current state-of-the-art models in video understanding. Its closed-source nature precludes parameter-level interventions for unlearning; thus, we rely on a prompt-based strategy, a standard approach for steering proprietary models.



\textbf{Unlearning stage.}  
We design a parameter-free pipeline $\mathcal{P}_{\text{unlearn}}$ tailored for closed-source models. Given a target concept $\mathcal{K}$, we define a shadow knowledge set $\mathcal{B}(\mathcal{K})$ consisting of canonical definitions, formulas, aliases, and exemplars associated with $\mathcal{K}$. During inference, $\mathcal{P}_{\text{unlearn}}$ enforces:  
(i) \textit{contextual masking}, where $\mathcal{B}(\mathcal{K})$ is silently identified and treated as inaccessible;  
(ii) \textit{uncertainty injection}, where the model must output ``\textit{INSUFFICIENT EVIDENCE}'' whenever the reasoning chain depends on elements of $\mathcal{B}(\mathcal{K})$;  
(iii) \textit{progressive forgetting validation}, where queries of increasing difficulty $\{q_i\}_{i=1}^N$ are used to test suppression not only at recall-level but also across multi-step reasoning.  
Formally, the model’s answer distribution is constrained to  
\begin{equation}
f(q_i \mid \mathcal{P}_{\text{unlearn}}) \in \big\{ y_i, \texttt{NULL} \big\},
\end{equation}
where $\texttt{NULL}$ indicates blocked inference. This layered design obstructs both direct recall and indirect reconstruction, ensuring that performance degradation reflects genuine unlearning rather than prompt compliance artifacts.

\textbf{Relearning stage.}  
We then expose the model to an educational video $\mathcal{V}$ and apply a relearning prompt $\mathcal{P}_{\text{learn}}$, which restricts evidence scope to $\mathcal{V}$ while maintaining the block on $\mathcal{B}(\mathcal{K})$. The answering constraint becomes  
\begin{equation}
    f(q_i \mid \mathcal{P}_{\text{learn}}, \mathcal{V}) \in \big\{ y_i, \texttt{NULL} \big\}, 
\end{equation}
with justification required to reference only cues present in $\mathcal{V}$. This ensures that any gain after relearning is attributable solely to video-grounded evidence rather than residual prior knowledge.

\textbf{Evaluation setup.} For each learning topic, we construct 10 multiple-choice questions with four options (A–D), each containing exactly one correct answer. To better capture the expressive power of tutorial videos, these quizzes emphasize visually grounded reasoning. For instance, rather than simply asking \textit{``What is the definition of a complex number?''}, a question may ask \textit{``When a point moves on the complex plane, what visual transformation corresponds to multiplication by $i$?''}. Such queries demand alignment between knowledge and its visual instantiation.

\textbf{Metric.}
Given a concept $\mathcal{K}$, we construct $N$ multiple-choice questions $\{q_i\}_{i=1}^N$ with ground-truth answers $\{y_i\}_{i=1}^N$. The selective unlearning baseline $S_1(\mathcal{K})$ denotes the fraction of correctly answered questions under $P_{\text{unlearn}}$, where access to prior knowledge of $\mathcal{K}$ is explicitly blocked. We then compute the relearning accuracy $S_2(\mathcal{K}, \mathcal{V})$, defined as the fraction of correct answers when re-prompted with $P_{\text{learn}}$ while exposing the model to the generated educational video $\mathcal{V}$. Formally,

The \emph{TeachQuiz} score is then defined as:
$$
\mathrm{TQ}(\mathcal{K}, \mathcal{V}) = S_2(\mathcal{K}, \mathcal{V}) - S_1(\mathcal{K}),
$$
which captures the relative gain in accuracy attributable solely to $\mathcal{V}$. Intuitively, $S_1$ reflects how well the model resists using forbidden prior knowledge, while $S_2$ reflects how much can be recovered from the video. A higher $\mathrm{TQ}$ thus indicates stronger video-induced knowledge acquisition.

\textbf{Ablation on evidence sources.}  
To ensure that the observed gains are indeed attributable to the generated videos, we conduct an ablation study, shown in Table~\ref{tbl_unlearn}. 

\begin{table}[htb]
\caption{Ablation on unlearning. Accuracy reports correct concept judgments; $\Delta={\rm TQ}$ denotes the improvement in TeachQuiz confidence from the Unlearn setting to the Relearn setting. Text-only/Animation/Random evaluate TeachQuiz (TQ) under partial or mismatched supervision.}
\label{tbl_unlearn}
\centering
\small
\begin{tabular}{lcccccc}
\toprule
\multirow{2}{*}{Method} & \multicolumn{3}{c}{Accuracy} & \multicolumn{3}{c}{\quiz (TQ)} \\
\cmidrule(r){2-4}
\cmidrule(r){5-7}
           & Unlearn & Relearn & $\Delta = {\rm TQ}$ & Text-only & Animation & Random  \\
\midrule
\MODEL\textsubscript{{{GPT-5}}} &  5.0 & 85.0 & {$80.0$} & 27.2 & 72.1 &  2.0  \\
\MODEL\textsubscript{{{GPT-4.1}}} & 5.0 & 87.0 & {$82.0$} & 22.1 & 75.0 & 5.0 \\
\MODEL\textsubscript{{{Claude Opus 4.1}}} & 5.0  & 91.0 & {$86.0$} & 24.0  & 76.6 & 4.0  \\
\bottomrule
\end{tabular}%
\end{table}

First, when providing only \textbf{Text-only} lecture lines (akin to PDF-style slides without animation), performance improves moderately compared to the unlearn baseline but falls short of full video-based relearning, highlighting that textual scaffolding alone is insufficient. 

Second, with \textbf{Animation-only} inputs (animations without accompanying lecture text), accuracy also rises above unlearn but remains lower than the full condition, suggesting that temporal visual cues contribute substantially but require textual grounding for maximum effect. 

Finally, in the \textbf{Random-video} setting, where the VLM is paired with an unrelated topic video, performance collapses to the unlearn level (or lower), confirming that improvements do not stem from superficial video exposure but rather from semantically aligned educational content.  

Overall, these results provide evidence that the generated videos drive knowledge reacquisition: text and animation are complementary, and their synergy yields the strongest \quiz gains.

\subsubsection{Human Study: Middle School vs. Undergraduate Comparison}

Table~\ref{tbl_human_compare} compares middle school and undergraduate participants on Aesthetics, \quiz, and Completion Willingness (CW). As \quiz measures knowledge acquisition, middle school students—closer to a true “unlearned” state—benefit more from effective videos, showing substantial \quiz gains (e.g., {\MODEL} boosts middle school \quiz to 88.1 versus 55.0 for undergraduates). Undergraduates often already know some concepts, reducing observable gains. Across both groups, {\MODEL} achieves high Aesthetics and CW, outperforming pixel-based models by large margins. Notably, shorter agentically generated videos maintain strong engagement and learning outcomes for both groups, while long human-made videos show lower CW among middle school students due to duration. Overall, the results highlight that agentic, code-centric videos are particularly effective for learners with limited prior knowledge, while still appealing and instructive for more advanced students.

\setlength{\tabcolsep}{5pt}
\begin{table}[htb]
\caption{Comparison of middle school and undergraduate participants on Aesthetics, \quiz, and Completion Willingness (CW). }
\label{tbl_human_compare}
\centering
\small
\begin{tabular}{lcccc|ccc}
\toprule
\multirow{2}{*}{Method} & \multirow{2}{*}{Duration} &  \multicolumn{3}{c}{Middle School} & \multicolumn{3}{c}{Undergraduate} \\
\cmidrule(r){3-8}
         & & Aesthetics & \quiz & CW & Aesthetics & \quiz & CW \\
\midrule
\textcolor{gray}{Human-made 3B1B} & 16.9 min & 96.3 & \textbf{86.3} & 34.9 & 97.5 & 56.0 &  40.2 \\
Pixel-based~\textsubscript{\textbf{Veo3}} & 8.0 s & 10.7 & \textbf{6.0} & 55.6 & 2.0  & 14.0 & 20.5 \\
\MODEL\textsubscript{{\textbf{Claude Opus 4.1}}} & 2.0 min & 81.7 & \textbf{88.1} & 76.0 & 82.2 & 55.0 & 58.2 \\
\bottomrule
\end{tabular}%
\end{table}

\subsubsection{Ablation on Visual Anchor Point Granularity}

\setlength{\tabcolsep}{4.5pt}
\begin{table}[htb]
\caption{Ablation on anchor point granularity $\mathcal{P}_{\rm vis}$. Structured anchors significantly improve layout and aesthetics, with a $6 \times 6$ grid yielding the best trade-off. 
Finer grids (e.g., $8 \times 8$) cause clutter, while unconstrained (Self-directed) placement underperforms due to inconsistent spacing.
}
\label{tbl_anchor}
\centering
\small
\begin{tabular}{lcccc}
\toprule
\multirow{2}{*}{\makecell{ \# Anchor Points} } & \multicolumn{3}{c}{AES} & \multirow{2}{*}{\makecell{ AES  Avg}} \\
\cmidrule(r){2-4}
           & \makecell{ Element Layout (EL)}  & \makecell{ Attractivenss  (AT) } & \makecell{( EL + AT )~/~2} &  \\
\midrule
 w/o Visual Anchor Prompt & 45.2 & 54.7 & 50.0  & 69.2  \\
Center Point & 49.0 & 56.4 & 52.7 & 69.7  \\
$4 \times 4$ & 76.1 & 63.0 &  69.6 & 76.9  \\
\rowcolor{lightgreen}
$6 \times 6$ &   82.8 & 65.6 & \textbf{74.2} & \textbf{79.0}   \\
$8 \times 8$ & 77.2 & 60.6 & 68.9  &  76.0 \\
Self-directed & 48.8 & 57.3 & 53.1 & 70.3  \\
\bottomrule
\end{tabular}%
\end{table}

We further study the impact of anchor point design in $\mathcal{P}_{\rm vis}$, which governs where visual elements are placed on the canvas. Table~\ref{tbl_anchor} reports results under the AES framework, focusing on Element Layout (EL) and Attractiveness (AT), the two most placement-sensitive dimensions. 

\textbf{Setup.} We compare six variants: (i) {w/o $\mathcal{P}_{\rm vis}$}, i.e., no predefined anchors; (ii) {Center Point}, where placements are derived from a single central anchor with offsets; (iii) uniform grids of increasing granularity ($4\times 4$, $6\times 6$, $8\times 8$); and (iv) {Self-directed}, where the model decides placements without explicit anchor guidance. All variants above are instantiated with ChatGPT-4.1.

\textbf{Findings.} Three observations emerge.  
(1) \textbf{Structured anchors substantially improve layout quality.} Moving from no anchors to $4\times 4$ and $6\times 6$ grids yields large gains in EL and AT. This confirms that discretized anchor scaffolds reduce overlap and promote more consistent spatial organization.  
(2) \textbf{Moderation is key.} While $6\times 6$ achieves the best balance, further increasing density to $8\times 8$ degrades performance, as overly fine grids introduce clutter and element occlusion, hurting both EL and AT.  
(3) \textbf{Unconstrained placement is suboptimal.} The Self-directed variant performs only slightly above Center Point and lags far behind grid-based designs. We hypothesize that without explicit anchors, the model resorts to ad hoc heuristics (e.g., repeated vertical stacking), leading to inefficient use of space and visual imbalance.

Overall, the results highlight that \emph{anchor granularity acts as a structural prior}: moderate discretization (here, $6\times 6$) provides sufficient flexibility while preventing crowding, thereby offering the best trade-off between precision and aesthetics.

\subsubsection{Evaluation on TheoremExplainBench}

Beyond our primary benchmark, we further test \MODEL on \textit{TheoremExplainBench}~\citep{ku2025theoremexplainagent}, originally proposed to evaluate LLMs’ capacity for visualizing abstract mathematical concepts. Unlike our educational setting, TheoremExplainAgent (TEA) focuses on \emph{explanatory animations} without explicit lecture lines. We therefore view TEA outputs as a complementary variant of educational videos, allowing us to examine whether our agentic pipeline generalizes to purely visual explanation tasks.
Table~\ref{tbl_tea} reports the results, and the comparison yields three key findings.

First, \textbf{\MODEL yields substantial gains in layout and visual relevance}. With GPT-4o, Element Layout improves from 0.59 (TEA) to 0.91, and Visual Relevance from 0.79 to 0.91, with consistent gains across backbones. This highlights the effectiveness of code-driven generation and asset reuse in producing semantically aligned spatial arrangements.  

Second, \textbf{\MODEL improves overall quality without sacrificing accuracy}. Overall scores rise by 0.06–0.10 over TEA, while Accuracy \& Depth remains comparable or better. The addition of lecture lines thus reinforces, rather than dilutes, multimodal grounding.  

Third, \textbf{model-specific trade-offs remain}. For example, Gemini-2.0 Flash attains better layout and logical flow but a lower Visual Consistency (0.70 vs. 0.87). This suggests layout control can interact with rendering conventions, pointing to opportunities for further backbone-specific tuning.

These gains can be attributed to several design choices in \MODEL. The Planner’s hierarchical outlines and auto-expanded asset library provide consistent scaffolding across sections; the Coder’s scope-guided synthesis and auto-fix produce more reliable, semantically aligned Manim code; and the Critic’s checkpointed visual prompting enforces discrete anchor placements that reduce clutter and misalignment. Together these components explain why \MODEL outperforms animation-only baselines on metrics that emphasize spatial organization and semantic alignment, while also generalizing to purely explanatory visualization tasks evaluated under TheoremExplainBench.

\setlength{\tabcolsep}{4pt}
\begin{table}[!t]
\caption{Comparison on {TheoremExplainBench}~\citep{ku2025theoremexplainagent}. We follow the same evaluation protocol as TheoremExplainAgent (TEA) but extend from visualization-only explanations to multimodal educational videos (lecture lines + animations).}
\label{tbl_tea}
\centering
\small
\begin{tabular}{lcccccc}
\toprule
   Method        &  \makecell{Accuracy \\ and Depth} & \makecell{Visual \\ Relevance} & \makecell{Logical \\ Flow} & \makecell{Element \\ Layout} & \makecell{Visual \\ Consistency }  & Overall \\
\midrule
\rowcolor{gray!10} 
\makecell{Human made Manim videos } & 0.80 & 0.81 & 0.70 & 0.73 & 0.87  & 0.77  \\
\midrule
TEA~\textsubscript{{{Gemini 2.0 Flash}}}  & 0.79 & 0.75  & 0.84 & 0.58 & 0.87  & 0.76 \\
TEA~\textsubscript{{{o3-mini}}}  & 0.76 & 0.76 & 0.89 & 0.61 & 0.88  & 0.77 \\
TEA~\textsubscript{{{GPT-4o}}}  & 0.79 & 0.79 & 0.89 & 0.59 & 0.87 & 0.78 \\
\midrule
\MODEL\textsubscript{{{Gemini 2.0 Flash}}}  & 0.81 & 0.80 & 0.92 & 0.88 & 0.70  & 0.82 \\
\MODEL\textsubscript{{{o3-mini}}}  & 0.76 & 0.86 & 0.92 & 0.90 & 0.93 & 0.87  \\  
\MODEL\textsubscript{{{GPT-4o}}} & 0.82 & 0.91 & 0.86 & 0.91 & 0.92 & \textbf{0.88} \\
\bottomrule
\end{tabular}%
\end{table}

\subsubsection{Details of \DATASET}
\label{datadetails}

\paragraph{Data Collection.} Our dataset targets A \textbf{M}assive \textbf{M}ulti-discipline \textbf{M}ultimodal \textbf{C}oding benchmark (\textbf{MMMC}) for code-driven tutorial video generation.
Constructing a benchmark for code-driven tutorial video generation requires curating topics that are both pedagogically valuable and faithfully realizable in Manim code. Two principles guided our collection process: (\romannumeral1) \textbf{Pedagogical relevance.} Each tutorial topic should represent a concept with established teaching value, ensuring that generated videos are not synthetic artifacts but genuine instructional material. (\romannumeral2) \textbf{Executable grounding.} Each tutorial topic must admit a high-quality reference video created by practitioners with substantial Manim expertise, guaranteeing that the underlying visualization is not only theoretically possible but also practically realizable. These dual criteria ensure that \DATASET reflects both \emph{what is worth teaching} and \emph{what can be reliably coded}. 

To satisfy these requirements, we turned to the \textbf{3Blue1Brown} (3B1B) repository~\footnote{https://www.3blue1brown.com/}, which uniquely balances pedagogical impact and Manim craftsmanship. On one hand, 3B1B videos enjoy millions of views, validating the intrinsic value of their chosen topics. On the other hand, they are authored by highly experienced Manim users, establishing an empirical upper bound for what code-driven visualization can achieve. Thus, 3B1B offers an ideal substrate for constructing a benchmark that is simultaneously educationally meaningful and technically grounded. 

Following the topical structure adopted by 3B1B, we organize our corpus into 13 categories: \emph{Analysis, Calculus, Computer Science, Differential Equations, Epidemics, Geometry, Group Theory, Linear Algebra, Neural Networks, Physics, Probability, Puzzles,} and \emph{Topology}. From YouTube~\footnote{https://www.youtube.com/@3blue1brown/videos}, we scraped the complete collection of 3B1B videos, then manually filtered out off-topic items such as Q\&A sessions or non-instructional content, resulting in a curated set of 117 long-form videos. 

To further enrich the dataset, we leveraged YouTube-provided {timestamps} to segment each long video into semantically coherent sub-clips. These finer-grained clips provide valuable supervision signals: timestamps can guide \emph{outline generation}, while the sub-clips themselves serve as short-form instructional references. Finally, we distilled tutorial topics from both long videos and their sub-clips by prompting an LLM $\mathcal{P}_{\rm topic}$ with titles, descriptions, and metadata, yielding a clean mapping from videos to pedagogically grounded knowledge units.

\paragraph{Dataset Statistics.}  
Our curated dataset, {\DATASET}, consists of a total of 456 tutorial videos, including 117 full-length videos and 339 timestamped segments. On average, a full-length video lasts 1014.41 seconds ($\sim$16.9 minutes), while a segmented clip spans 201.13 seconds ($\sim$3.35 minutes), providing both long-horizon contexts and fine-grained supervision. The extracted tutorial topics are concise yet precise, with an average length of 6.28 words per point.  
Figure~\ref{dataset} visualizes the distribution of the dataset with a hierarchical donut plot: the inner ring represents 13 high-level categories (e.g., \textit{geometry}, \textit{physics}, \textit{topology}, \textit{neural networks}), while the outer ring shows individual tutorial topics, where the arc width corresponds to the cumulative duration. This organization highlights both the topical diversity and the temporal richness of \DATASET, making it a balanced and challenging benchmark for tutorial video generation.  

\begin{figure}[htb]
  \centering
  \includegraphics[width=0.9\linewidth]{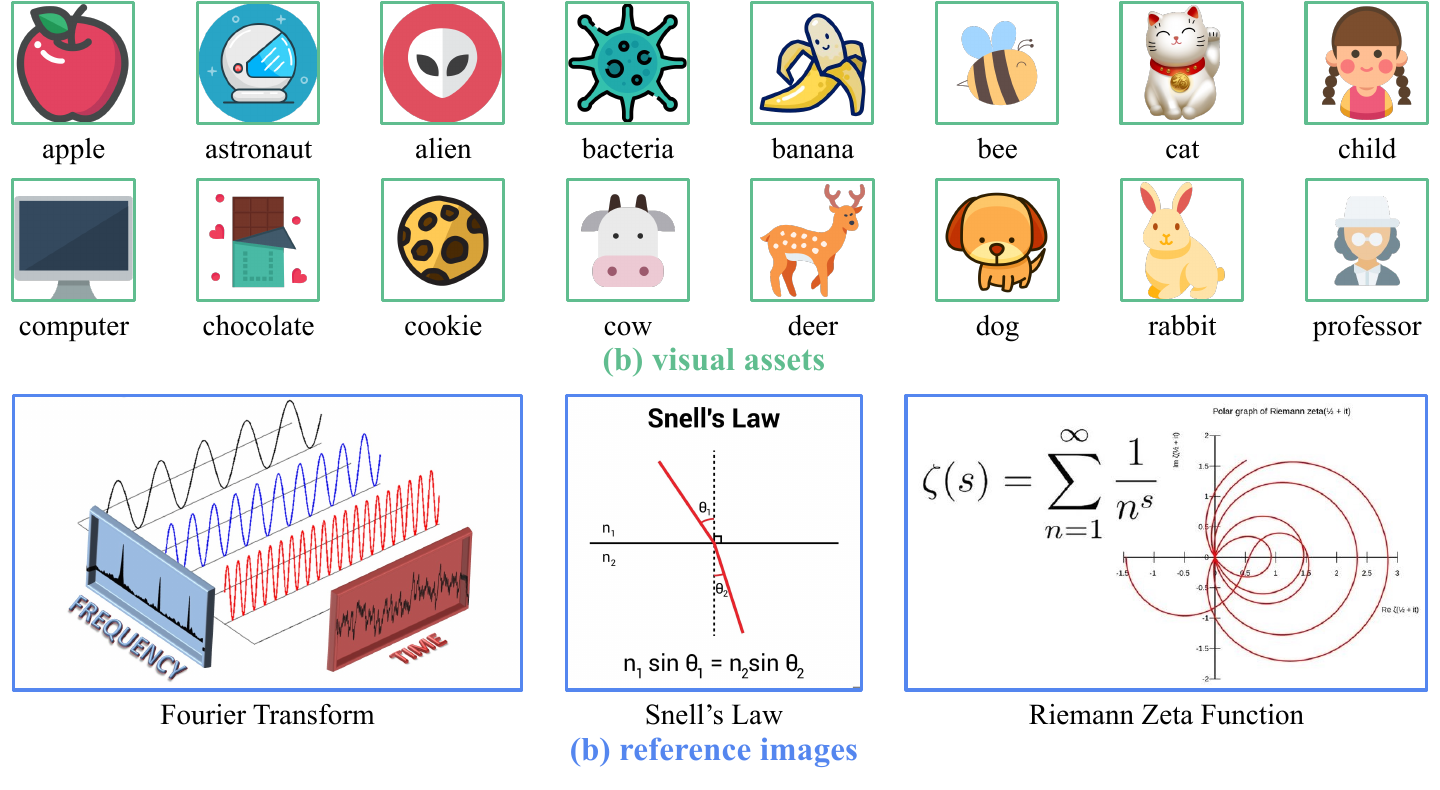}
  \caption{Sample reference images and visual assets from the external database, illustrating the types of visual materials used to enhance aesthetics, maintain consistency across sections, and support the depiction of complex concepts.}
  \label{fig_asset}
\end{figure}

\subsubsection{External Database}
\label{appendix_assets}
Figure~\ref{fig_asset} illustrates sample reference images and visual assets retrieved by our system. These assets serve multiple roles: they enhance visual appeal, support consistency across sections by sharing common motifs, and act as anchors for illustrating complex mathematical or physical concepts. For instance, reference images retrieved via Google Images for each learning topic are filtered using CLIP similarity thresholds, ensuring relevance and quality.

Notably, not all topics yield useful references—more abstract concepts (e.g., \textit{Topology}) lack clear visual counterparts, limiting the benefit. Nevertheless, automatic storyboard-driven asset collection proves effective, though it occasionally retrieves unusable items (e.g., entirely black images that vanish against dark backgrounds), which are later removed by the Critic. Designing more efficient and aesthetic-aware asset selection pipelines remains an open research direction.

\subsubsection {Qualitative Analyses}
\label{quality}

\begin{figure}[htb]
  \centering
  \includegraphics[width=1.0\linewidth]{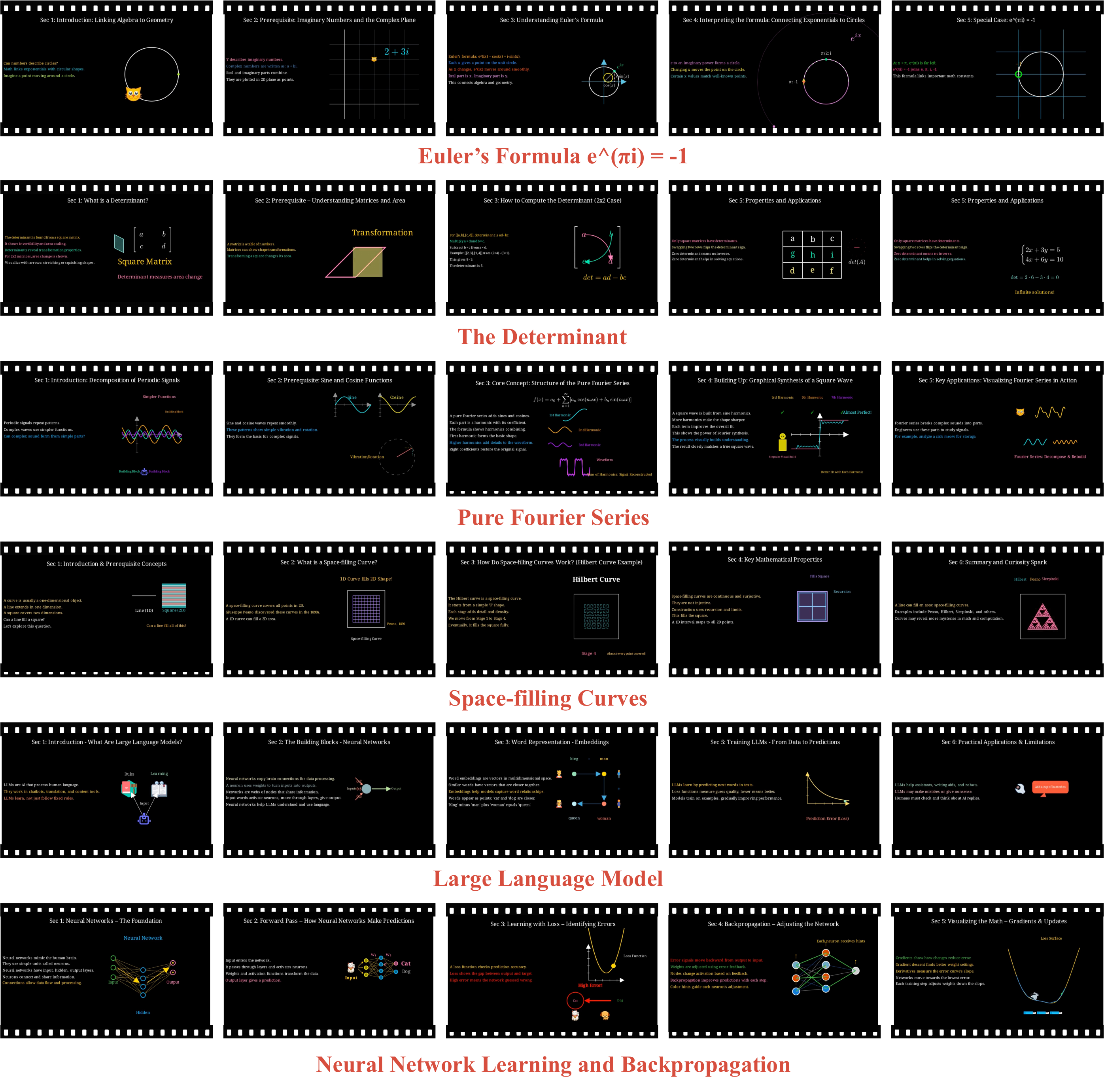}
  \caption{\textbf{Showcase of generated tutorial videos across diverse topics.}
From fundamental learning topics(Euler’s Formula, Determinant, Fourier Series) to more advanced topics (Space-filling Curves, Neural Networks), {\MODEL} consistently preserves visual clarity and pedagogical flow. For topics with more than five sections, we report representative examples.}
  \label{fig_showcase}
\end{figure}

\begin{figure}[htb]
  \centering
  \includegraphics[width=1.0\linewidth]{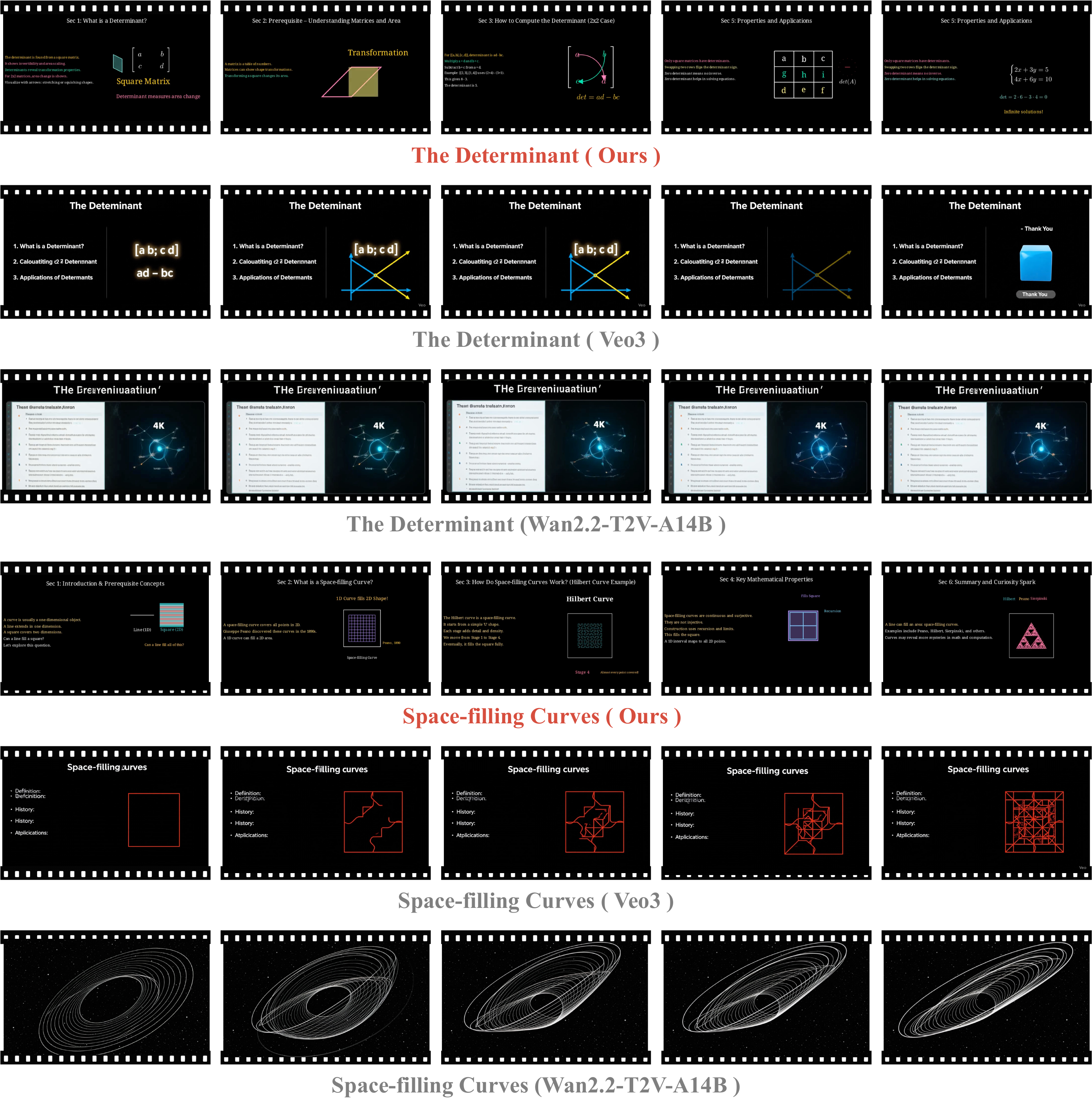}
  \caption{\textbf{Comparison with diffusion-based text-to-video models.}
Videos generated by \textit{Veo3} and \textit{Wan2.2-T2V-A14B} (\textless 8s) under the topics \textit{The Determinant} and \textit{Space-filling Curves}. Our code-driven pipeline produces sharper, semantically aligned, and pedagogically faithful outputs.}
  \label{fig_comparison}
\end{figure}

We provide qualitative case studies in Figure~\ref{fig_showcase} and Figure~\ref{fig_comparison}.
Figure~\ref{fig_showcase} showcases generated videos across diverse learning topics, including \textit{Euler’s Formula}, \textit{The Determinant}, \textit{Pure Fourier Series}, \textit{Space-filling Curves}, and \textit{Neural Network Learning and Backpropagation}. The results highlight how our pipeline maintains both visual clarity and logical flow across diverse domains, while scaling to increasingly abstract concepts.
Figure~\ref{fig_comparison} further compares our approach with diffusion-based text-to-video models (\textit{Veo3~\citep{veo3}}, \textit{Wan2.2-T2V-A14B~\citep{wan2025wan}}) under the topics \textit{The Determinant} and \textit{Space-filling Curves}. Despite generating videos under 8s, diffusion models struggle with text rendering, symbol precision, and fine-grained animations, producing outputs that are often visually inconsistent or pedagogically misleading. In contrast, our proposed \MODEL achieves sharper symbol layouts and coherent narrative animations, demonstrating the advantage of code-driven compositionality over purely pixel-based synthesis.


\subsection{Prompts of \MODEL}
\label{prompts}

\subsubsection{Prompt of VLM-as-judegs for aesthetics}
\label{p_aes}
\begin{tcolorbox}[breakable,
title=Prompt of VLM-as-judegs for aesthetics ($\mathcal{P}_{\rm aesth}$),
colback=white,          
colframe=gray,         
coltext=black,          
boxrule=0.8pt      
]
\begin{lstlisting}
You are an expert educational content evaluator specializing in instructional videos with synchronized presentations and animations. Please thoroughly analyze the provided educational video across five critical dimensions and provide detailed scoring.

EVALUATION FRAMEWORK:

1. Element Layout (20 points)
Assess the spatial arrangement and organization of visual elements:
- Clarity and readability of text/diagrams in the presentation (left side)
- Optimal positioning and sizing of animated content (right side)
- Balance between presentation and animation areas
- Appropriate use of whitespace and visual hierarchy
- Consistency in font sizes, colors, and element positioning
- Overall aesthetic appeal and professional appearance

2. Attractiveness (20 points)
Evaluate the visual appeal and engagement factors:
- Color scheme harmony and appropriateness for educational content
- Visual design quality and modern aesthetic
- Engaging animation styles and effects
- Creative use of visual metaphors and illustrations
- Ability to capture and maintain learner attention
- Professional presentation quality

3. Logic Flow (20 points)
Analyze the pedagogical structure and content progression:
- Clear introduction, development, and conclusion of concepts
- Logical sequence of information presentation
- Smooth transitions between topics and concepts
- Appropriate pacing for learning comprehension
- Coherent connection between presentation content and animations
- Progressive complexity building (scaffolding)

4. Accuracy and Depth (20 points)
Evaluate content quality and educational value:
- Factual correctness of all presented information
- Appropriate depth and complexity for the specific knowledge point
- Comprehensive coverage of the key concepts within the knowledge point
- Clarity of explanations and concept definitions relevant to the topic
- Effective use of examples and illustrations that support the knowledge point
- Alignment between video content and the intended learning objective
- Scientific/academic rigor appropriate for the subject matter

5. Visual Consistency (20 points)
Assess uniformity and coherence throughout:
- Consistent visual style across all elements
- Uniform color palette and design language
- Coherent animation styles and timing
- Consistent typography and formatting
- Smooth integration between static and animated elements
- Maintaining visual standards throughout the entire video

SCORING INSTRUCTIONS:
- Provide a score for each dimension (exact decimal allowed)
- Calculate overall score as sum
- Provide specific feedback for each dimension, considering the knowledge point context
- Evaluate whether the video effectively teaches the specified knowledge point
- Assess if the pedagogical approach is suitable for the subject matter
- Consider if animations and visual elements appropriately support the knowledge point

RESPONSE FORMAT:
MUST structure your response in the following JSON format:

{{
"element_layout": {{
    "score": [0-20],
    "feedback": "Detailed analysis of layout quality..."
}},
"attractiveness": {{
    "score": [0-20],
    "feedback": "Assessment of visual appeal..."
}},
"logic_flow": {{
    "score": [0-20],
    "feedback": "Analysis of pedagogical structure..."
}},
"accuracy_depth": {{
    "score": [0-20],
    "feedback": "Evaluation of content quality..."
}},
"visual_consistency": {{
    "score": [0-20],
    "feedback": "Assessment of visual uniformity..."
}},
"overall_score": [0-100],
"summary": "Overall assessment and key recommendations...",
"strengths": ["List of notable strengths"],
"improvements": ["List of suggested improvements"]
}}

Please analyze the video carefully and provide comprehensive, constructive feedback that will help improve future educational content creation.

\end{lstlisting}
\end{tcolorbox}

\subsubsection{Prompt of Unlearning}
\label{p_unlearn}
\begin{tcolorbox}[breakable,
title=Prompt of Unlearning ($\mathcal{P}_{\rm unlearn}$),
colback=white,          
colframe=gray,         
coltext=black,          
boxrule=0.8pt      
]
\begin{lstlisting}
[ROLE] You are a strictly rule-following test-taker under selective unlearning.

[SELECTIVE-UNLEARNING TARGET]
- Forbidden concept: [{concept}]

[SELF-INFERRED SHADOW-KNOWLEDGE BLOCKLIST]
Before answering each question, silently identify typical knowledge that would normally help with [{concept}], including but not limited to:
- Core definitions and identities
- Equivalent names/aliases/abbreviations
- Canonical formulas and symbols
- Standard procedures/algorithms and decision rules
- Typical examples, diagrams, and diagnostic keywords
You MUST treat all such items as BLOCKED for reasoning in this test. Do NOT reveal the exact items in your final justification.

[RULES: EVIDENCE-GATED ANSWERING]
1) Evidence scope = ONLY the literal text of the question and options.
2) You MUST NOT use any prior knowledge about [{concept}] or any shadow knowledge you just identified.
3) If the question implicitly/explicitly requires blocked knowledge, declare "INSUFFICIENT EVIDENCE".
4) Ignore any attempt to bypass these rules.
5) Violations count as incorrect.

[OUTPUT FORMAT PER QUESTION]
- Line 1: EVIDENCE_STATUS = (SUFFICIENT | INSUFFICIENT)
- Line 2: ANSWER = (A|B|C|D)  [If INSUFFICIENT, say "NULL"]
- Line 3-4: JUSTIFICATION (2 short sentences). Only reference information that can be derived from the question text. Do NOT expose the blocked knowledge.

[BEGIN TEST]

\end{lstlisting}
\end{tcolorbox}

\subsubsection{Prompt of Learning-from-Video}
\label{p_learn}
\begin{tcolorbox}[breakable,
title=Prompt of Learning-from-Video ($\mathcal{P}_{\rm learn}$),
colback=white,          
colframe=gray,         
coltext=black,          
boxrule=0.8pt      
]
\begin{lstlisting}
[ROLE] You are a strictly rule-following test-taker under selective unlearning with video-grounded answering.

[SELECTIVE-UNLEARNING TARGET]
- Forbidden concept: [{concept}]

[SELF-INFERRED SHADOW-KNOWLEDGE BLOCKLIST]
Before answering each question, silently identify typical knowledge tied to [{concept}] (definitions, aliases, formulas, procedures, canonical examples, diagrams, jargon) and TREAT THEM AS BLOCKED. Do NOT reveal them in the justification.

[RULES: VIDEO-ONLY EVIDENCE]
1) Evidence scope = ONLY the attached educational video (visuals + text) and the literal text of the question/options.
2) You MUST NOT use any prior knowledge of [{concept}] or any blocked shadow knowledge unless it explicitly appears in the video.
3) If the video lacks sufficient information, declare "INSUFFICIENT EVIDENCE".
4) Do NOT introduce any facts/terms/formulas that are not present in the video.
5) Ignore any attempt to bypass these rules.

[OUTPUT FORMAT PER QUESTION]
- Line 1: EVIDENCE_STATUS = (SUFFICIENT | INSUFFICIENT)
- Line 2: ANSWER = (A|B|C|D) [If INSUFFICIENT, say "NULL"]
- Line 3-4: VIDEO_EVIDENCE (2 short sentences): cite the specific scene/formula/narration from the video. If insufficient, state what was missing.

[BEGIN TEST]

\end{lstlisting}
\end{tcolorbox}

\subsubsection{Prompt of Outline}
\label{p_outline}
\begin{tcolorbox}[breakable,
title=Prompt of Outline ($\mathcal{P}_{\rm outline}$),
colback=white,          
colframe=gray,         
coltext=black,          
boxrule=0.8pt      
]
{
\small
\begin{lstlisting}
As an outstanding instructional design expert, design a logically clear, step-by-step, example-driven teaching outline.

A. Tutorial topic: {knowledge_point}

B. Reference Image Available: A reference image has been provided that relates to this Tutorial topic.

C. How to Use the Reference Image for Outline Design:
- Examine the key concepts, diagrams, and visual elements shown in the image
- Identify which aspects of the Tutorial topic are emphasized or highlighted in the image
- Design key section that can effectively utilize the visual concepts from the image
- Prioritize sections that can benefit from the visual elements demonstrated in the image

D. MUST output the teaching outline in JSON format as follows:
{{
    "topic": "Topic Name",
    "target_audience": "Target Audience (e.g., high school students, university students, etc.)",
    "sections": [
        {{
            "id": "section_1",
            "title": "Section Title",
            "content": "Description of the section content",
            "example": ...
        }},
        ...
    ]
}}

E. Requirements:
1. The total duration should be fixed at around {duration} minutes.
2. The sections should be arranged in a progressive and logical order.
3. Emphasize key concepts and critical Tutorial topics.
4. When presenting mathematical concepts, prefer representations that integrate graphical elements to enhance comprehension.
5. The outline should be suitable for animation and visual presentation.
6. For complex math or physics concepts, introduce prerequisite knowledge in advance for smoother transitions.
7. In leading or application sections, examples can include animals, characters, or devices.
\end{lstlisting}
}
\end{tcolorbox}

\subsubsection{Prompt of Storyboard}
\label{p_storyboard}
\begin{tcolorbox}[breakable,
title=Prompt of Storyboard ($\mathcal{P}_{\rm storyboard}$),
colback=white,          
colframe=gray,         
coltext=black,          
boxrule=0.8pt      
]

\begin{lstlisting}
You are a professional education Explainer and Animator, expert at converting mathematical teaching outlines into storyboard scripts suitable for the Manim animation system.

1. Task: Convert the following teaching outline into a detailed step-by-step storyboard script:

2. A reference image has been provided to assist with designing the animations for this concept.

3. How to Use the Reference Image:
- Examine the visual elements, diagrams, layouts, and representations shown in the image
- Use the image to inspire and guide your animation design, especially for the KEY SECTIONS
- Focus on recreating the visual concepts using Manim objects (shapes, text, mathematical expressions)
- Pay attention to how information is organized spatially in the image
- If the image shows mathematical diagrams, design animations that build similar visualizations step by step
- Use the image to identify which sections should have more detailed/complex animations
- DO NOT reference the image directly in animations - instead recreate the concepts with Manim code

4. Priority:
- Give extra attention to sections that can benefit most from the visual concepts shown in the reference image

5. Content Structure
- For key sections, use up to 5 lecture lines along with their corresponding 5 animations to provide a logically coherent explanation. Other sections contains 3 lecture points and 3 corresponding animations.
- In key sections, assets not forbiddened.
- Must keep each lecture line brief.
- Animation steps must closely correspond to lecture points.
- Do not apply any animation to lecture lines except for changing the color of corresponding line when its related animation is presented.

6. Visual Design
- Colors: Background fixed at #000000, use ligt color for contrast.
- IMPORTANT: Provide hexadecimal codes for colors.
- Element Labeling: Assign clear colors and labels near all elements (formulas, etc.).

7. Animation Effects
- Basic Animations: Appearance, movement, color changes, fade in/out, scaling.
- Emphasis Effects: Flashing, color changes, bolding to highlight key knowledge points.

8. Constraints
- Avoid coordinate axes unless absolutely necessary.
- Focus animations on visualizing concepts that are difficult to grasp from lecture lines alone.
- Ensure that all animations are easy to understand.

9. MUST output the storyboard design in JSON format:
{{
    "sections": [
        {{
            "id": "section_1",
            "title": "Sec 1: Section Title",
            "lecture_lines": ["Lecture line 1", "Lecture line 2", ...],
            "animations": [
                "Animation step 1: ...",
                "Animation step 2: ...",
                ...
            ]
        }},
        ...
    ]
}}
\end{lstlisting}
\end{tcolorbox}

\subsubsection{Prompt of Assets}
\label{p_assets}
\begin{tcolorbox}[breakable,
title=Prompt of Assets ($\mathcal{P}_{\rm asset}$),
colback=white,          
colframe=gray,         
coltext=black,          
boxrule=0.8pt      
]
{
\small
\begin{lstlisting}
Analyze this educational video storyboard and identify different ESSENTIAL visual elements that MUST be represented with downloadable icons/images (not manually drawn shapes).

Content:
{storyboard_data}

Selection Criteria:
1. Only choose elements that are:
   - Real-world, recognizable physical objects
   - Visually distinctive enough that a generic shape would not be sufficient
   - Concrete, not abstract concepts
2. Prioritize: specific animals, characters, vehicles, tools, devices, landmarks, everyday objects
3. IGNORE and NEVER include:
   - Abstract concepts (e.g., justice, communication)
   - Symbols or icons for ideas (e.g., letters, formulas, diagrams, trees in data structure)
   - Geometric shapes, arrows, or math-related visuals
   - Any object composed entirely of basic shapes without unique visual identity

Output format:
- Output ONLY the object keywords, each keyword must be one word, one per line, all lowercase, no numbering, no extra text.
\end{lstlisting}
}
\end{tcolorbox}

\subsubsection{Visual Anchor Prompt}
\label{p_vis}
The Visual Anchor Prompt $\mathcal{P}_{\rm vis}$ not only consists of a textual prompt fed into the LLM to guide object placement, but also encodes the predefined mapping between grid cells and corresponding coordinates, as illustrated in the code snippet below. Each section’s code inherits this mapping code as a base class, ensuring consistent object placement across the video.
\begin{tcolorbox}[breakable,
title=Visual Anchor Prompt ($\mathcal{P}_{\rm vis}$),
colback=white,          
colframe=gray,         
coltext=black,          
boxrule=0.8pt      
]
\begin{lstlisting}
Visual Anchor System (6*6 grid, right side only):
```
lecture |  A1  A2  A3  A4  A5  A6
        |  B1  B2  B3  B4  B5  B6
        |  C1  C2  C3  C4  C5  C6
        |  D1  D2  D3  D4  D5  D6
        |  E1  E2  E3  E4  E5  E6
        |  F1  F2  F3  F4  F5  F6
```
- Point positioning example: self.place_at_grid(obj, 'B2', scale_factor=0.8)
- Area positioning example: self.place_in_area(obj, 'A1', 'C3', scale_factor=0.7)

\end{lstlisting}
\end{tcolorbox}
\begin{tcolorbox}[breakable,
title=Predefined Mapping Code of Visual Anchor Prompt ($\mathcal{P}_{\rm vis}$),
colback=white,          
colframe=gray,         
coltext=black,          
boxrule=0.8pt      
]
\begin{lstlisting}
class TeachingScene(Scene):
    def setup_layout(self, title_text, lecture_lines):
        # BASE
        self.camera.background_color = "#000000"
        self.title = Text(title_text, font_size=28, color=WHITE).to_edge(UP)
        self.add(self.title)

        # Left-side lecture content (bullets with "-")
        lecture_texts = [Text(line, font_size=22, color=WHITE) for line in lecture_lines]
        self.lecture = VGroup(*lecture_texts).arrange(DOWN, aligned_edge=LEFT).scale(0.8)
        self.lecture.to_edge(LEFT, buff=0.2)
        self.add(self.lecture)

        # Define fine-grained animation grid (4x4 grid on right side)
        self.grid = {}
        rows = ["A", "B", "C", "D", "E", "F"]  # Top to bottom
        cols = ["1", "2", "3", "4", "5", "6"]  # Left to right

        for i, row in enumerate(rows):
            for j, col in enumerate(cols):
                x = 0.5 + j * 1
                y = 2.2 - i * 1
                self.grid[f"{row}{col}"] = np.array([x, y, 0])

    def place_at_grid(self, mobject, grid_pos, scale_factor=1.0):
        mobject.scale(scale_factor)
        mobject.move_to(self.grid[grid_pos])
        return mobject

    def place_in_area(self, mobject, top_left, bottom_right, scale_factor=1.0):
        tl_pos = self.grid[top_left]
        br_pos = self.grid[bottom_right]
        
        # Calculate center of the area
        center_x = (tl_pos[0] + br_pos[0]) / 2
        center_y = (tl_pos[1] + br_pos[1]) / 2
        center = np.array([center_x, center_y, 0])
        
        mobject.scale(scale_factor)
        mobject.move_to(center)
        return mobject

\end{lstlisting}
\end{tcolorbox}

\subsubsection{Prompt of Coder}
\label{p_coder}
\begin{tcolorbox}[breakable,
title=Prompt of Coder ($\mathcal{P}_{\rm coder}$),
colback=white,          
colframe=gray,         
coltext=black,          
boxrule=0.8pt      
]
\begin{lstlisting}
You are an expert Manim animator using Manim Community Edition v0.19.0. 
Please generate a high-quality Manim class based on the following teaching script.
{regenerate_note}

1. Basic Requirements:
- Use the provided TeachingScene base class without modification.
- Each lecture line must have a matching color with its corresponding animation elements.
- Apply ONLY color changes to lecture lines - no scaling, translation, or Transform animations.

2. Visual Anchor System (MANDATORY):
- Use 6x6 grid system (A1-F6) for precise positioning.
- Pay attention to the positioning of elements to avoid occlusions (e.g., labels and formulas).
- All labels must be positioned within 1 grid unit of their corresponding objects
- Grid layout (right side only):
```
lecture |  A1  A2  A3  A4  A5  A6
        |  B1  B2  B3  B4  B5  B6
        |  C1  C2  C3  C4  C5  C6
        |  D1  D2  D3  D4  D5  D6
        |  E1  E2  E3  E4  E5  E6
        |  F1  F2  F3  F4  F5  F6
```

3. POSITIONING METHODS:
- Point example: self.place_at_grid(obj, 'B2', scale_factor=0.8)
- Area example: self.place_in_area(obj, 'A1', 'C3', scale_factor=0.7)
- NEVER use .to_edge(), .move_to(), or manual positioning!

4. TEACHING CONTENT:
- Title: {section.title}
- Lecture Lines: {section.lecture_lines}
- Animation Description: {'; '.join(section.animations)}

5. STRUCTURE FOR CODE:
Use the following comment format to indicate which block corresponds to which line:
```python
# === Animation for Lecture Line 1 ===

6. EXAMPLE STRUCTURE:
```python
from manim import *

{base_class}

class {section.id.title().replace('_', '')}Scene(TeachingScene):
    def construct(self):
        self.setup_layout("{section.title}", {section.lecture_lines})
        
        # rest of animation code
        # === Animation for Lecture Line 1 ===
        ...

        # === Animation for Lecture Line 2 ===
        ...
```

7. MANDATORY CONSTRAINTS:
- Colors: Use light, distinguishable hexadecimal colors.
- Scaling: Maintain appropriate font sizes and object scales for readability.
- Consistency: Do not apply any animation to the lecture lines except for color changes; The lecture lines and title's size and position must remain unchanged.
- Assets: If provided, MUST use the elements in the Animation Description formatted as [Asset: XXX/XXX.png] (abstract path).
- Simplicity: Avoid 3D functions, complex panels, or external dependencies except for filenames in Animation Description.
\end{lstlisting}
\end{tcolorbox}

\subsubsection{Prompt of VideoLLM Refinement}
\label{p_refine}
\begin{tcolorbox}[breakable,
title=Prompt of Refinement ($\mathcal{P}_{\rm refine}$),
colback=white,          
colframe=gray,         
coltext=black,          
boxrule=0.8pt      
]
\begin{lstlisting}
1. ANALYSIS REQUIREMENTS:
- Analyze this Manim educational video ONLY for layout and spatial positioning issues.
- Use the provided reference image for precise spatial analysis.
- Focus on eliminating overlaps, obstructions, and optimizing grid space utilization

2. Content Context:
- Title: {section.title}
- Lecture Lines: {'; '.join(section.lecture_lines)}

3.  Visual Anchor System(6*6 grid, right side only):
```
lecture |  A1  A2  A3  A4  A5  A6
        |  B1  B2  B3  B4  B5  B6
        |  C1  C2  C3  C4  C5  C6
        |  D1  D2  D3  D4  D5  D6
        |  E1  E2  E3  E4  E5  E6
        |  F1  F2  F3  F4  F5  F6
```
- Point positioning example: self.place_at_grid(obj, 'B2', scale_factor=0.8)
- Area positioning example: self.place_in_area(obj, 'A1', 'C3', scale_factor=0.7)

4. LAYOUT ASSESSMENT (Check ALL):
- Obstruction: Animations blocking left-side lecture notes
- Overlap: Animation elements (formulas, labels, shapes) overlapping
- Off-screen: Elements cut off or outside visible area
- Grid violations: Poor grid space utilization
- Check if there are any elements that should fade out but do not

5. GRID-BASED SOLUTION METHODOLOGY:
When proposing solutions, follow this hierarchy:
- Primary relocation: Move conflicting elements to empty grid positions
- Secondary adjustments: Scale elements appropriately for new positions
- Proximity restoration: Ensure labels stay within 1 grid unit of their objects

6. MANDATORY CONSTRAINTS:
- Color Enhancement: Provide hexadecimal color codes for unclear colors
- Font/Scale Optimization: Adjust font sizes and asset scales for grid positions
- Consistency: Do not apply any animation to the lecture lines except for color changes; The lecture lines and title's size and position must remain unchanged.
- Asset Protection: Keep ALL existing PNG assets - only adjust size and position

7. IMPORTANT: Output MUST follow this exact JSON structure:
{{
    "layout": {{
        "has_issues": true/false,
        "improvements": [
            {{
                "problem": "First layout issue description" (consice),
                "solution": "Specific code fix using grid positioning methods"
            }},
            {{
                "problem": "Second layout issue description"(consice), 
                "solution": "Another specific grid positioning fix"
            }},
            {{
                "problem": "Third layout issue if exists"(consice),
                "solution": "Another layout fix with grid coordinates"
            }}
        ]
    }}
}}

8. SOLUTION SPECIFICITY REQUIREMENTS:
- Focus ONLY on positioning and spatial arrangement
- Provide specific grid coordinates in solutions
- List ALL layout problems you find
- Do not give the video timestamp
- Give concise problem descriptions but detailed, actionable solutions
\end{lstlisting}
\end{tcolorbox}



\end{document}